\documentclass[sigconf, screen, nonacm]{acmart}
\AtBeginDocument{%
  }

\usepackage{algorithm}
\usepackage{algorithmic}
\usepackage{multirow}

\usepackage{wrapfig}
\usepackage{caption}

\begin{document}

\title{SpatialGrammar: A Domain-Specific Language for LLM-Based 3D Indoor Scene Generation}

\author{Song Tang}
\affiliation{%
  \institution{The Hong Kong University of Science and Technology (Guangzhou)}
  \country{}
}

\author{Kaiyong Zhao}
\affiliation{%
  \institution{XGRIDS}
  \country{}
}

\author{Yuliang Li}
\affiliation{%
  \institution{XGRIDS}
  \country{}
}

\author{Qingsong Yan}
\affiliation{%
  \institution{XGRIDS}
  \country{}
}

\author{Penglei Sun}
\affiliation{%
  \institution{The Hong Kong University of Science and Technology (Guangzhou)}
  \country{}
}

\author{Junyi Zou}
\affiliation{%
  \institution{XGRIDS}
  \country{}
}

\author{Qiang Wang}
\authornote{Corresponding author.}
\affiliation{%
  \institution{Harbin Institute of Technology (Shenzhen)}
  \country{}
}

\author{Xiaowen Chu}
\authornotemark[1]
\affiliation{%
  \institution{The Hong Kong University of Science and Technology (Guangzhou)}
  \country{}
}

\renewcommand{\shortauthors}{Tang et al.}

\begin{abstract}
    Automatically generating interactive 3D indoor scenes from natural language is crucial for virtual reality, gaming, and embodied AI. However, existing LLM-based approaches often suffer from spatial errors and collisions, in part because common scene representations—raw coordinates or verbose code—are difficult for models to reason about 3D spatial relationships and physical constraints. We propose SpatialGrammar, a domain-specific language that represents gravity-aligned indoor layouts as BEV grid placements with deterministic compilation to valid 3D geometry, enabling verifiable constraint checking. Building on this representation, we develop (1) SG-Agent, a closed-loop system that uses compiler feedback to iteratively refine scenes and enforce collision constraints, and (2) SG-Mini, a 104M-parameter model trained entirely on compiler-validated synthetic data. Across 159 test scenes spanning five scenarios of different complexity, SG-Agent improves spatial fidelity and physical plausibility over prior methods, while SG-Mini performs competitively against larger LLM-based baselines on single-shot generation scenarios. Our project page is available at \url{https://xgrids-3d.github.io/SpatialGrammar/}
\end{abstract}


\settopmatter{authorsperrow=3}
\maketitle

  \section{Introduction}

  The automated generation of interactive 3D indoor scenes from natural language is increasingly important for virtual reality, gaming, and embodied AI, where agents require diverse, physically grounded environments for training and evaluation~\cite{savva2019habitat,xia2018gibson}. Compared with generating a single object or a static rendering, scene generation must produce a coherent environment with many objects, explicit identities, and controllable spatial relationships—properties essential for procedural manipulation, reuse, and physical simulation~\cite{yang2024holodeck,hu2024scenecraft}.

  Despite recent progress, LLM-based structured scene generation still suffers from spatial inaccuracies and collisions, and often struggles with complex natural-language constraints. A major reason is representational: common scene formats are either too geometric (e.g., raw 6-DoF coordinates and poses) or too verbose (e.g., long code or deeply nested structures), making it difficult for models to reason reliably about 3D spatial relationships and physical constraints. Existing approaches—whether directly predicting coordinates~\cite{gu2025artiscene,yang2024llplace} or generating intermediate representations like JSON or Python~\cite{yang2024holodeck,hu2024scenecraft,bucher2025respace}—remain either spatially unintuitive or unreliable when post-processing must enforce geometric validity. This raises a central question: can we design a representation that is both spatially intuitive for LLMs and reliably executable under physical constraints?

  We address this question by encoding strong geometric and physical priors directly into the representation, rather than forcing LLMs to internalize full 3D physics. Inspired by Bird's-Eye View (BEV) representations in autonomous driving~\cite{yang2023bevformer,li2023bevstereo,wei2023surroundocc}, we observe that indoor environments share a similar structure: gravity aligns most objects with a common ground plane, and many layout decisions are naturally expressed in 2D. Based on this observation, we propose SpatialGrammar, a domain-specific language that reframes 6-DoF pose generation as intuitive BEV grid placement with gravity-aligned orientation, while supporting hierarchical sub-layouts (e.g., tabletop arrangements, shelves, chessboards) through face-anchored local frames—the same mechanism also enables wall-mounted objects and ceiling attachments by anchoring grids to vertical or overhead faces. A companion architectural DSL specifies floorplans as wall segments with openings. Crucially, SpatialGrammar is deterministically compilable into valid 3D geometry, enabling verifiable constraint checking (e.g., collisions and support) during generation. This turns scene creation into iterative program editing: the model operates on compact DSL text while the compiler handles geometric details and provides actionable feedback.

  \begin{figure}[!t]
      \centering
      \includegraphics[width=0.9\linewidth]{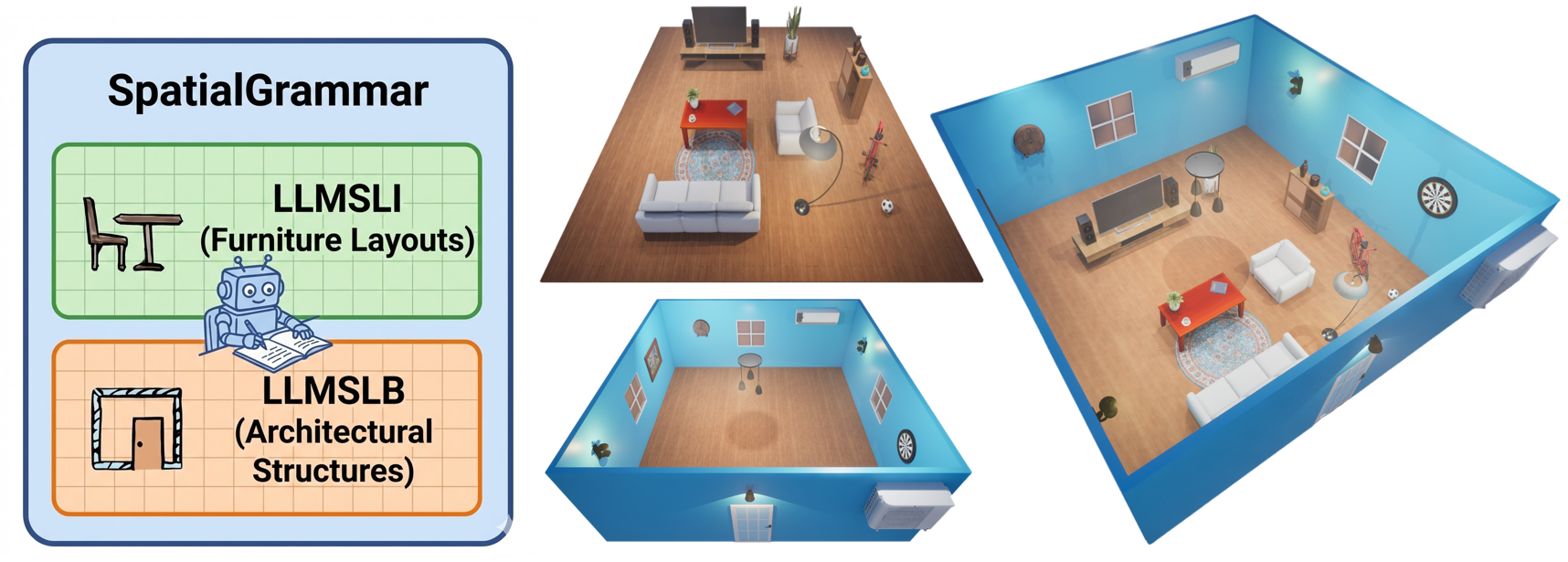}
      \caption{The LLM agent generates complete 3D indoor scenes, including furniture layout and wall structure, by writing in our proposed SpatialGrammar DSL.}
      \label{fig:teaser}
      \vspace{-15pt}
  \end{figure}

  This representation has two practical and research-relevant implications. First, because compilation provides explicit, structured feedback about constraint violations (e.g., "object A overlaps object B"), we can close the loop between generation and execution: we instantiate this as SG-Agent, a tool-augmented system that iteratively proposes and refines SpatialGrammar programs using compiler feedback and rendering-based scene feedback, improving reliability when satisfying spatial constraints and user edits. Second, the same compiler that executes SpatialGrammar also serves as a scalable validator for data synthesis. This makes it possible to train a lightweight model without human annotation or reliance on closed-source frontier LLMs: we construct a fully synthetic pipeline and train SG-Mini, a 104M-parameter small language model specialized for SpatialGrammar generation.

  In summary, we make three contributions. (1) We introduce \textit{SpatialGrammar}, a deterministically compilable domain-specific language that represents indoor layouts as BEV grid placements, supporting hierarchical arrangements while remaining spatially intuitive. (2) We develop \textit{SG-Agent}, a closed-loop system that leverages compiler feedback to improve constraint satisfaction and scene quality under natural-language instructions. (3) We present a compiler-validated synthetic training pipeline and demonstrate \textit{SG-Mini}, a 104M-parameter model that performs competitively against larger LLM-based baselines on single-shot generation scenarios, validating the model-friendliness of our DSL design.

  \section{Related Work}
  \label{sec:related}

  Text-to-3D scene generation broadly follows two directions: implicit methods based on neural rendering and diffusion~\cite{DreamFusion,metzer2023latent,
  mildenhall2021nerf,Text2Room} prioritize visual fidelity but typically entangle semantics and geometry in implicit fields, making object-level identities, editability, and physical verification difficult.

  In contrast, explicit and structured methods aim to produce interactive environments with manipulable objects. Some works directly predict continuous object poses from language~\cite{yang2024llplace}, which can make it difficult to reliably satisfy discrete spatial constraints and avoid collisions without explicit verification. Other pipelines use visual intermediates or search to infer 3D placements, e.g., by generating 2D images and lifting to 3D~\cite{gu2025artiscene} or by hierarchical VLM-guided search~\cite{deng2025global}. Another line generates intermediate structured specifications---such as scene graphs, JSON, or executable programs---to improve modularity and enable editing~\cite{yang2024holodeck,ocal2024sceneteller,bucher2025respace,feng2023layoutgpt,pun2025hsm}. SceneCraft~\cite{hu2024scenecraft} generates Blender code with VLM-based feedback for iterative refinement. DirectLayout~\cite{ran2025direct} uses BEV layout as an intermediate representation before lifting to 3D, sharing the insight that BEV abstraction benefits spatial reasoning, though without hierarchical sub-layouts or a compilable DSL for formal verification. OptiScene~\cite{yang2025optiscene} trains scene generation models via SFT and DPO with injected spatial violations as hard negatives, a strategy conceptually related to our Error Chain method; our approach differs by chaining multiple error types (semantic, spatial, collision, and syntax) to produce compounding failure patterns that more closely mimic real model failures. Despite these advances, most representations remain either spatially unintuitive or token-inefficient, and many pipelines still require additional post-processing to enforce geometric validity.

  Recent work also explores richer settings: AnyHome~\cite{fu2024anyhome} and Open-Universe~\cite{aguina2024openuniverse} target open-vocabulary generation with diverse asset databases; Procedural Scene Programs~\cite{gumin2025proceduralsceneprograms} study automated error correction via program search; FirePlace~\cite{huang2025fireplace} and SpatialLM~\cite{mao2025spatiallm} combine language models with geometric constraints for structured indoor understanding. Across these lines of work, a recurring challenge is to simultaneously achieve (i) \textit{deterministic executability} and verifiable physical validity, (ii) \textit{fine-grained controllability and editability} for complex constraints, and (iii) \textit{spatially intuitive representations} that are token-efficient for language models.

  \section{Methodology}

  \begin{figure}[!t]
      \centering
      \includegraphics[width=0.9\linewidth]{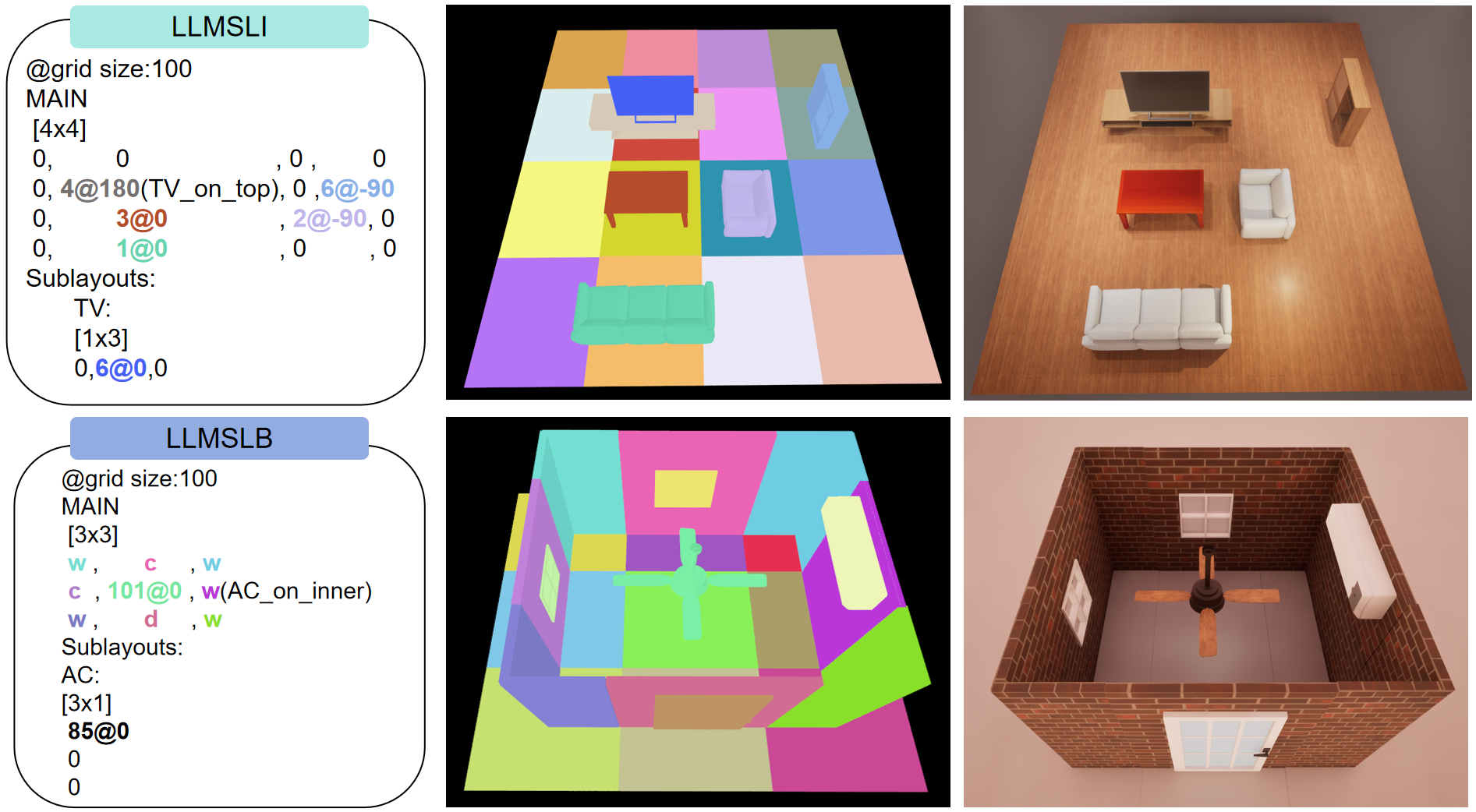}
      \caption{Each row shows DSL code (left), semantic intermediate representation (middle), and final render (right). Top: LLMSLI places furniture on a BEV grid with a sub-layout attaching a TV onto the stand. Bottom: LLMSLB defines walls and openings, with a sub-layout mounting an air conditioner on the wall.}
      \label{fig:dsl_mechanism}
      \vspace{-10pt}
  \end{figure}

  \begin{figure}[!t]
      \centering
      \includegraphics[width=0.9\linewidth]{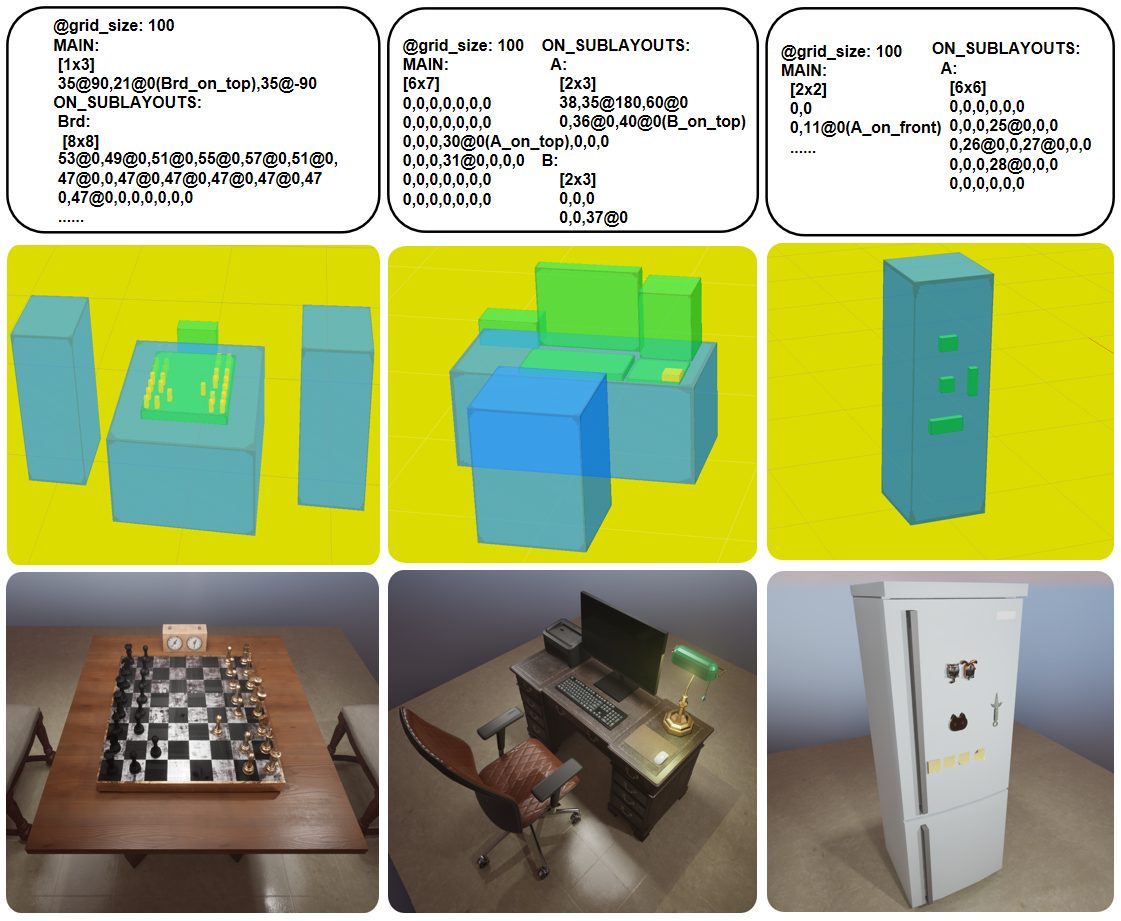}
      \caption{Sub-layout examples for hierarchical scene control. Each column shows (top to bottom) the LLMSLI content, semantic 3D box layout, and final rendered scene for three scenarios: a chessboard with pieces, an office desk with supplies, and a refrigerator with items on its front door.}
      \label{fig:sublayout}
      \vspace{-10pt}
  \end{figure}

  \begin{figure}[!t]
      \centering
      \includegraphics[width=0.9\linewidth]{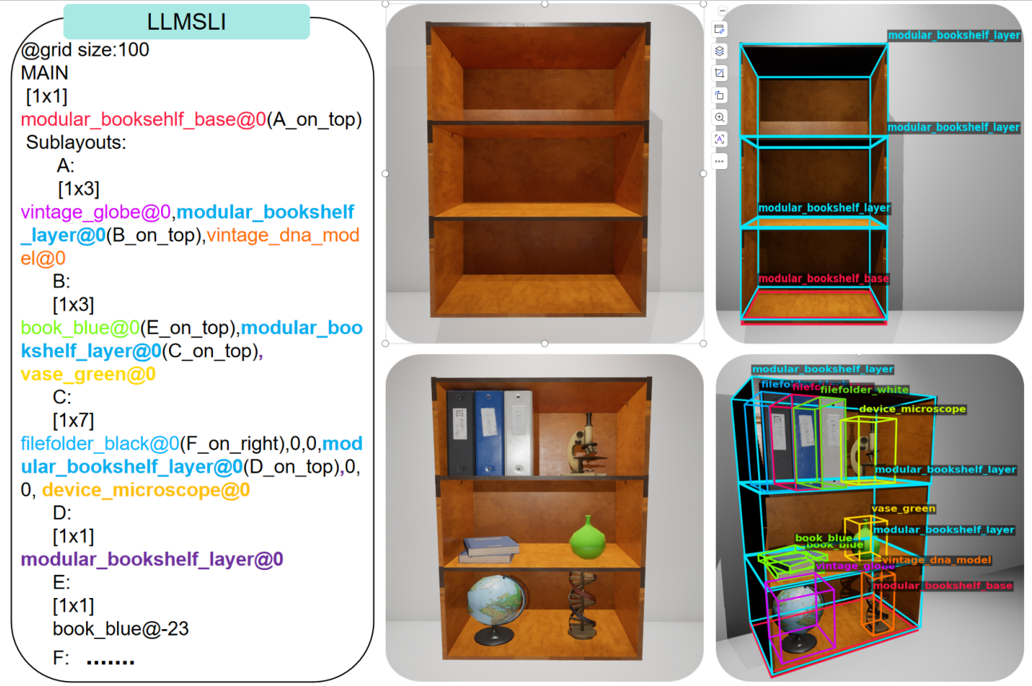}
      \caption{Modular asset composition based on our sub-layout system. Using descriptive string identifiers to retrieve assets from external asset databases, the sub-layout system composes a multi-layered bookshelf with items placed on each shelf surface.}
      \label{fig:modular_assemble}
      \vspace{-10pt}
  \end{figure}

  Our methodology centers on \textit{SpatialGrammar}, a Domain-Specific Language that reduces cognitive load for 3D scene generation by encoding physical priors into the representation. We first present the language design (\S\ref{sec:language}), demonstrating how Bird's-Eye View abstraction and recursive sub-layouts transform complex 3D coordinate generation into intuitive 2D text generation. Building upon this foundation, we develop an agentic workflow (\S\ref{sec:agent}) that realizes conversational 3D scene creation, enabling an AI agent to interpret natural language, iteratively refine layouts through reliable closed-loop feedback, and output validated scenes. Finally, we propose a self-sufficient training pipeline (\S\ref{sec:slm_training}) that exploits the language's structured nature to automatically synthesize training data through a complete three-stage process—pre-training, supervised fine-tuning (SFT), and direct preference optimization (DPO)—enabling specialized Small Language Models to master this task without human annotation.

  \subsection{The SpatialGrammar Language}
  \label{sec:language}

  SpatialGrammar comprises two tightly coupled DSLs: LLMSLI (Large Language Model Spatial Layout Indoor) for furniture and object arrangements, and LLMSLB (Large Language Model Spatial Layout Building) for architectural structures. LLMSLB extends the same design principles and abstractions used in LLMSLI to the building envelope, while LLMSLI remains the core component for indoor scene layout. In this subsection, we therefore first detail LLMSLI, showing how its Bird's-Eye View (BEV) design turns 3D scene specification into intuitive 2D symbolic layout that LLMs can easily generate.

  \subsubsection{BEV-Based Representation}

  The core insight of LLMSLI is \textit{dimensional reduction}: rather than forcing LLMs to reason about full 3D coordinates $(x, y, z, \text{pitch}, \text{yaw}, \text{roll})$, we abstract scenes into a 2D Bird's-Eye View (BEV) grid. This reduces the problem from 6-DOF pose generation to intuitive 2D spatial layout, where the LLM specifies object placements on a grid analogous to arranging furniture on a floor plan. In the root BEV grid, objects are implicitly placed on the floor, so the LLM never has to predict metric heights directly. More complex height variations, such as objects on tables or shelves, are handled by the sub-layout system and discussed in detail in \S\ref{sec:sublayouts}. The compiler then performs the geometric conversion: given integer $(row, col)$ indices and a global grid size $g$ defined in the LLMSLI header, it maps each cell to planar coordinates $(x, y) = (i \cdot g, j \cdot g)$ and uses the object's 3D box dimensions to place its center in $z$ such that the bottom face exactly touches the supporting surface. The model therefore only predicts discrete grid indices and yaw angles, while the compiler recovers precise 3D poses. This design embeds physical priors about gravity, support surfaces, and permissible rotations, and greatly reduces cognitive load without sacrificing expressiveness: a living room layout that would require hundreds of tokens in JSON format with explicit $(x, y, z)$ coordinates becomes a compact 2D grid where spatial relationships are immediately visible (see Figure~\ref{fig:dsl_mechanism}, top row, for a concrete example).

  \subsubsection{Compact Expression for Object Control}

  Within the BEV grid established above, each object is specified through a compact syntax that controls three essential properties: semantics, pose, and size. \textbf{Semantics} support two retrieval modes. A \textit{numerical mode} maps integer codes to object types through a curated vocabulary (e.g., \texttt{1} for sofa, \texttt{2} for coffee table), keeping the grid representation terse and well-suited for SLM training. An \textit{open-vocabulary mode} uses descriptive string identifiers (e.g., vintage\_globe, table\_round\_wood) to retrieve assets from external databases such as PolyHaven and Unreal Engine Fab, preserving the LLM's open-vocabulary strengths for richer, style-specific scene authoring. Both modes share the same compilation pipeline; the identifier is simply resolved to a 3D asset with calibrated default dimensions. Figure~\ref{fig:modular_assemble} demonstrates the open-vocabulary mode in a complex bookshelf scenario. \textbf{Pose} is controlled via the \texttt{@yaw} notation, where the LLM specifies an integer yaw angle in degrees (e.g., \texttt{@0}, \texttt{@39}, \texttt{@156}); any angle is valid, not only axis-aligned multiples. \textbf{Size} can optionally be specified via \texttt{[L×W×H]} notation, but most objects use sensible defaults from the vocabulary, avoiding unnecessary numbers in typical scenes. Figure~\ref{fig:dsl_mechanism} illustrates this object-level control, where a concise LLMSLI snippet (e.g., \texttt{4@180(TV\_on\_top)}) is compiled into a full 3D scene layout. Formally, each cell in the BEV grid is represented as:
  \begin{equation}
  \label{eq:object_dsl}
  o = 
  \begin{cases}
  (\text{id}, i, j, \theta) \in \mathcal{V} \times \mathbb{Z}^2 \times [0, 2\pi) & \text{if occupied} \\
  0 & \text{if empty}
  \end{cases}
  \end{equation}
  where $\text{id} \in \mathcal{V}$ is the furniture code, $(i,j)$ are discrete grid indices, and $\theta$ is the yaw angle. For occupied cells, the compiler $\mathcal{C}$ deterministically maps to a 3D bounding box:
  \begin{equation}
  \label{eq:object_compile}
  \mathcal{C}(o) = \mathbf{B} = 
  \begin{bmatrix}
  \mathbf{c} \\
  \mathbf{s} \\
  \mathbf{R}
  \end{bmatrix}
  =
  \begin{bmatrix}
  (i \cdot g, \; j \cdot g, \; h(\text{id})) \\
  \mathbf{s}(\text{id}) \\
  \text{Rot}_z(\theta)
  \end{bmatrix}
  \end{equation}
  where $g$ is the grid cell size, $h(\text{id})$ computes vertical placement based on object type, $\mathbf{s}(\text{id})$ denotes the default 3D box size retrieved from the object vocabulary, and $\text{Rot}_z(\theta)$ represents yaw-only rotation. Empty cells ($o = 0$) are ignored during compilation, allowing flexible spatial layouts.

  \subsubsection{Sub-Layouts for Complex Expression}
  \label{sec:sublayouts}
  Real-world indoor scenes contain not just individual objects but rich hierarchies of objects on top of other objects, such as office desks covered with supplies, chessboards densely filled with pieces, and refrigerator doors lined with condiments. LLMSLI captures these structures through \textit{recursive sub-layouts} that extend the BEV grid concept hierarchically. The primary mechanism is the \textbf{ON} sub-layout: for any parent object, we attach a local BEV grid to one of its faces and place child objects on that grid. In the main layout, this is triggered by annotations such as \texttt{62@30(B\_on\_top)}, where \texttt{B\_on\_top} denotes a sub-layout B anchored on the top face of the object with code \texttt{62}; other faces are addressed analogously via \texttt{\_on\_bottom}, \texttt{\_on\_left}, \texttt{\_on\_right}, \texttt{\_on\_front}, and \texttt{\_on\_back}. Most everyday relations in indoor scenes are expressed in this way, including objects on office desks, chessboards, and refrigerator doors, as illustrated by the \texttt{TV\_on\_top} example in Figure~\ref{fig:dsl_mechanism} and the more complex ON sub-layout examples in Figure~\ref{fig:sublayout}. Throughout, the LLM only reasons about local grids, while the compiler composes these local frames into global 3D poses. For a nesting chain $o_1 \supset o_2 \supset \cdots \supset o_k$, where $o_1$ is the root object in world space and $o_i$ (for $i > 1$) are specified in their respective parent's local frame, the deepest child's world-space bounding box is:

  \begin{equation}
  \label{eq:recursive_transform}
  \mathbf{B}_k^{\text{world}} = \mathbf{B}_1 \oplus \mathbf{B}_2^{\text{local}} \oplus \cdots \oplus \mathbf{B}_k^{\text{local}}
  \end{equation}
  where the transformation operator $\oplus$ composes parent and child frames as:
  \begin{equation}
  \label{eq:frame_composition}
  \mathbf{B}_p \oplus \mathbf{B}_c^{\text{local}} = 
  \begin{bmatrix}
  \mathbf{c}_c^{\text{world}} \\
  \mathbf{s}_c \\
  \mathbf{R}_c^{\text{world}}
  \end{bmatrix}
  =
  \begin{bmatrix}
  \mathbf{c}_p + \mathbf{R}_p \cdot \mathbf{c}_c^{\text{local}} \\
  \mathbf{s}_c \\
  \mathbf{R}_p \cdot \mathbf{R}_c^{\text{local}}
  \end{bmatrix}
  \end{equation}
  Here, each $\mathbf{B}_i^{\text{local}}$ is compiled from the DSL specification using Eq.~\ref{eq:object_compile}, with the local grid scaled to fit the parent's surface, and the operator $\oplus$ simply accumulates centers and orientations along the nesting chain to propagate local sub-layouts into consistent world-space coordinates. Notably, by combining modular assets---e.g., decomposing a bookshelf into a base unit and stackable shelf layers---the ON sub-layout mechanism naturally extends to ``inside'' arrangements such as placing items on individual shelf compartments, without requiring a separate containment primitive. Figure~\ref{fig:modular_assemble} demonstrates this: each \texttt{modular\_bookshelf\_layer} serves as a parent surface for the objects placed upon it, enabling fine-grained control over multi-layered interior structures through the same recursive composition.

  \subsubsection{Extension to Architectural Structures}

  While the LLMSLI language described above handles furniture and object arrangements, complete scene generation also requires architectural elements such as walls, openings, and ceilings. We extend \textit{SpatialGrammar} with LLMSLB (Large Language Model Spatial Layout Building), which follows the same BEV-based design philosophy but adapts vocabulary and syntax for structural elements. As shown in Figure~\ref{fig:dsl_mechanism} (bottom row), LLMSLB specifies walls via symbol codes (e.g., \texttt{w} for wall, \texttt{d} for door, \texttt{c} for window) on a BEV grid, with openings placed along wall edges. Face-anchored sub-layouts such as \texttt{AC\_on\_inner} enable wall-mounted objects like air conditioners. This architectural extension maintains the same cognitive simplicity of designing a floor plan on a 2D grid, while the compiler automatically generates proper 3D geometry. The unified BEV abstraction across both LLMSLI and LLMSLB enables end-to-end scene generation, from architectural shells to furnished interiors, all within a coherent framework.

  \subsection{Agent-Driven Scene Creation and Refinement}
  \label{sec:agent}

\begin{figure*}[!t]
      \centering
      \includegraphics[width=0.92\textwidth]{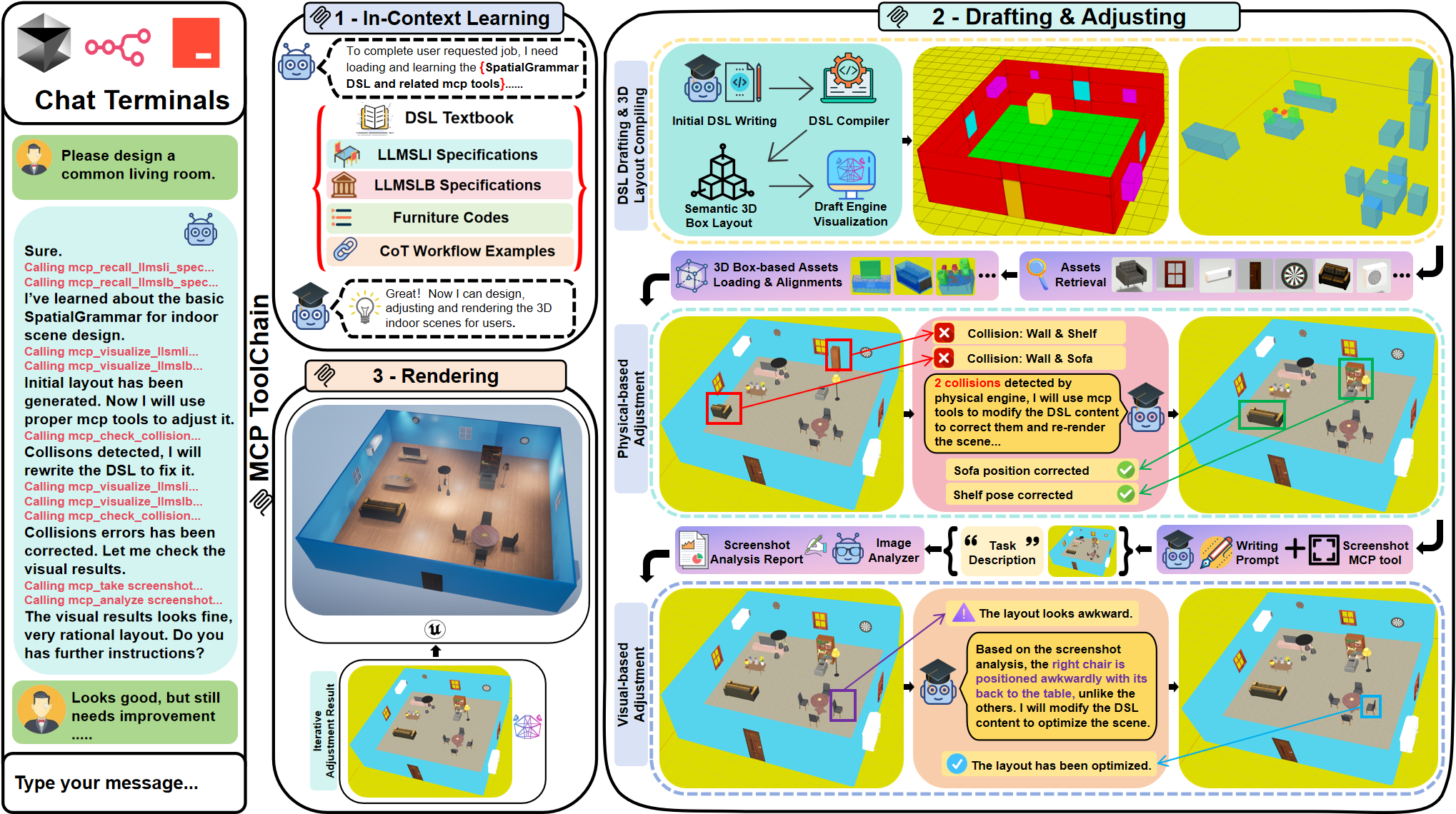}
      \caption{Overview of the SpatialGrammar agentic workflow. A user issues a request in the chat terminal, then the agent uses MCP tools to consult the DSL textbook and specifications, writes an initial SpatialGrammar script, and calls the compiler to obtain a semantic 3D box layout. After asset retrieval, this layout is loaded into the Draft Engine, which provides real-time visualization, physics-based collision checks, and screenshots for multimodal analysis. The agent iteratively edits the DSL based on symbolic and visual feedback until the scene is satisfactory, then exports the validated script to professional engines such as Unreal Engine for high-fidelity rendering.}
      \label{fig:method_overview}
  \end{figure*}

  Having established \textit{SpatialGrammar} as a compact and expressive representation, we now build a complete agentic workflow for conversational 3D scene generation (Figure~\ref{fig:method_overview}), implemented by a tool-augmented assistant that we refer to as \textit{SG-Agent}. The language's structured nature enables reliable translation to 3D geometry, while its spatial intuitiveness supports iterative refinement through multimodal feedback, yielding a system that integrates smoothly with existing design tools and downstream applications.

  \subsubsection{Reliable Scene Generation}

  The core of our workflow is the \textit{SpatialGrammar Compiler}, a deterministic rule-based translator that converts DSL text into precise 3D bounding boxes with 6-DOF poses. Grid indices map to exact $(x, y)$ coordinates, yaw angles to rotation matrices, and sub-layouts recursively inherit parent transformations, so any syntactically valid program yields a geometrically valid scene. This eliminates unreliability common in prior LLM-based pipelines, where layout outputs must pass through learned decoders or optimization stages that can introduce additional errors. The compiled boxes are rendered in the \textit{Draft Engine}, a lightweight 3D sandbox built on Ursina Engine that provides real-time visualization, physics-based collision checks, and screenshot capture for multimodal feedback. The engine supports two rendering modes: a \textit{semantic box view} for debugging and collision analysis, and an \textit{asset-mapped view} with retrieved 3D models for visual assessment. The agent can treat it as ground truth and focus on high-level spatial reasoning rather than low-level coordinate arithmetic.

  \subsubsection{Closed-Loop Refinement}

  The Draft Engine enables a closed-loop refinement mechanism through multimodal feedback. After each generation step, the agent receives two types of feedback: \textbf{symbolic feedback} from collision detection (e.g., "Sofa overlaps with coffee table at position (3,4)") and \textbf{visual feedback} from rendered screenshots analyzed by the agent. Symbolic signals provide precise, actionable diagnostics that guide targeted code corrections, while the agent (equipped with vision-language capabilities) interprets rendered images to assess global aesthetics, functional layout coherence (e.g., whether a sofa faces the TV), and spatial balance—aspects that are difficult to express symbolically. The agent follows a ReAct-style~\cite{yao2022react} iterative loop, reasoning about user requirements and current feedback and then acting by modifying the SpatialGrammar code, until collision-free and visually satisfactory results are achieved. Because the compiler and Draft Engine are deterministic and reliable, this loop converges in only a few iterations (typically 2–3 even for complex scenes). Importantly, the Draft Engine serves as a \textit{verification oracle}—it detects and reports collisions but does not automatically reposition objects; all layout corrections are made by the agent editing the DSL code.

  \subsubsection{Collaborative Scene Editing}

  Our workflow is orchestrated through the Model Context Protocol (MCP)~\cite{mcp}, which exposes the Draft Engine's capabilities as callable tools, enabling human-in-the-loop editing where users refine scenes through natural language while the agent updates the DSL and visualizes changes in real time. A dual-mode asset pipeline maps either numerical codes or open-vocabulary string identifiers to 3D models from curated libraries or external databases (PolyHaven, Unreal Engine Fab) for downstream rendering. As shown in Figure~\ref{fig:sublayout}, \textit{SG-Agent} can precisely place named chess pieces on specific squares in response to natural-language commands (e.g., ``put a white queen on d4''), demonstrating fine-grained controllability over both spatial layout and object identity.

  \subsection{Synthetic Data based SLMs Training}
  \label{sec:slm_training}

  While commercial LLMs (e.g., GPT-4, Claude) can generate \textit{SpatialGrammar} through few-shot prompting, their computational expense and closed-source nature hinder widespread adoption. We instead train specialized Small Language Models (SLMs) that match or exceed commercial model performance on our tasks, while using fewer parameters and achieving faster inference. In particular, we focus on a 104M-parameter SLM, \textit{SG-Mini}, which we describe in detail below. Two properties of \textit{SpatialGrammar} make such self-sufficient training possible: it is \textbf{model-friendly}, with intuitive BEV representations, structured syntax, and token-efficient encoding that simplify learning, and \textbf{compiler-friendly}, with a deterministic compiler that provides automatic correctness checks and supports programmatic generation of diverse, validated training data. Building on these properties, we construct a three-stage training pipeline consisting of pre-training, supervised fine-tuning (SFT), and direct preference optimization (DPO)~\cite{rafailov2023direct}, entirely from synthetic data and without human annotation. Algorithm~\ref{alg:training_pipeline} outlines the complete workflow.

  \begin{algorithm}[t]
  \caption{Self-Sufficient Training Pipeline}
  \label{alg:training_pipeline}
  \begin{algorithmic}[1]
  \REQUIRE Scene logic definitions $\mathcal{S}$, NL template library $\mathcal{T}$, base model $\pi_0$
  \ENSURE Trained SLM $\pi^*$

  \STATE \textbf{// Step 1: Programmatic SFT Data Generation}
  \STATE $\mathcal{D}_{\text{sft}} \gets \emptyset$
  \FOR{each scene logic specification $c \in \mathcal{S}$}
      \STATE Generate DSL code $s$, prompt $x$, and reasoning $r$ from $c$ and templates $\mathcal{T}$
      \IF{$\mathcal{C}(s) \neq \bot$ \AND $\text{CollisionFree}(s)$}
          \STATE $\mathcal{D}_{\text{sft}} \gets \mathcal{D}_{\text{sft}} \cup \{(x, r, s)\}$
      \ENDIF
  \ENDFOR

  \STATE \textbf{// Step 2: Pre-training Data Extraction}
  \STATE $\mathcal{D}_{\text{pre}} \gets \{x, r, s \mid (x,r,s) \in \mathcal{D}_{\text{sft}}\}$

  \STATE \textbf{// Stage 1: Pre-training on mixed corpus}
  \STATE $\pi_{\text{pre}} \gets \textsc{PreTrain}(\pi_0, \mathcal{D}_{\text{pre}})$

  \STATE \textbf{// Stage 2: SFT on (prompt, code) pairs only}
  \STATE $\mathcal{D}_{\text{sft-pairs}} \gets \{(x, s) \mid (x,r,s) \in \mathcal{D}_{\text{sft}}\}$ \COMMENT{Exclude $r$}
  \STATE $\pi_{\text{sft}} \gets \textsc{SFT}(\pi_{\text{pre}}, \mathcal{D}_{\text{sft-pairs}})$

  \STATE \textbf{// Step 3: Generate DPO pairs via Error Chain}
  \STATE $\mathcal{D}_{\text{dpo}} \gets \emptyset$
  \FOR{each $(x, s_w)$ in $\mathcal{D}_{\text{sft-pairs}}$}
      \STATE $s_l \gets \textsc{ErrorChain}(s_w)$
      \STATE $\mathcal{D}_{\text{dpo}} \gets \mathcal{D}_{\text{dpo}} \cup \{(x, s_w, s_l)\}$
  \ENDFOR

  \STATE \textbf{// Stage 3: DPO on preference pairs}
  \STATE $\pi^* \gets \textsc{DPO}(\pi_{\text{sft}}, \mathcal{D}_{\text{dpo}})$

  \RETURN $\pi^*$
  \end{algorithmic}
  \end{algorithm}

  \subsubsection{Self-Sufficient Training}

  We first construct a synthetic data engine that exploits SpatialGrammar's structured nature to programmatically generate training samples. It defines parameterized templates for common scene types (living rooms, bedrooms, offices) with configurable constraints on object counts and spatial relationships, then systematically varies these parameters and samples from furniture vocabularies to produce diverse scene descriptions paired with corresponding SpatialGrammar code. The compiler serves as an automatic correctness oracle: every generated program is validated by compilation and collision checking, ensuring the training set contains only valid, physically plausible scenes and avoiding the noise and inconsistencies of human-annotated datasets. For each scene, we also synthesize reasoning traces that mirror expert spatial reasoning, decomposing user requests into subgoals and justifying object placements. These reasoning traces are included in the pre-training corpus as supplementary language modeling data, exposing the model to spatial reasoning patterns during next-token prediction, but are \textit{not} used as supervision targets in SFT or DPO---the model learns to generate code directly from prompts without producing explicit chain-of-thought at inference time, keeping output compact for the 104M-parameter budget. For a single living room scenario, the engine can generate tens of thousands of pre-training samples and a few thousand SFT samples within minutes, demonstrating the pipeline's scalability.

  \subsubsection{Error Chain Method}

  Direct Preference Optimization (DPO)~\cite{rafailov2023direct} requires paired samples of preferred and dispreferred responses to the same prompt. Given a prompt $x$ (e.g., "design a cozy living room") and paired responses, a valid scene $y_w$ and a rejection $y_l$, we optimize the model to prefer $y_w$ over $y_l$:
  \begin{equation}
  \label{eq:dpo_objective}
  \resizebox{0.95\linewidth}{!}{$
  \mathcal{L}(\pi_\theta) = -\mathbb{E}_{(x,y_w,y_l) \sim \mathcal{D}_{\text{ErrorChain}}} \left[ \log \sigma \left( \beta \log \frac{\pi_\theta(y_w|x)}{\pi_\theta(y_l|x)} \right) \right]
  $}
  \end{equation}
  where $\beta$ controls the strength of preference, and $\mathcal{D}_{\text{ErrorChain}}$ denotes the dataset generated by our Error Chain method. The quality of rejections $y_l$ is critical: while generating high-quality chosen samples is straightforward using our validated synthetic data, creating \textit{meaningful} rejected samples, i.e., responses that are plausible but subtly flawed, poses a significant challenge. Random corruptions produce trivially bad outputs that provide weak learning signals; we need rejections that capture realistic failure modes. We introduce the \textbf{Error Chain} method to systematically generate such samples. Starting from a valid scene, we apply a sequence of controlled error injections: (1) \textit{semantic errors} (e.g., replacing a sofa with a random incompatible object like a bathtub in a living room), (2) \textit{spatial relationship errors} (e.g., placing a TV behind a sofa instead of in front), (3) \textit{collision errors} (e.g., overlapping furniture positions), and (4) \textit{syntax errors} (e.g., malformed grid coordinates or missing delimiters). Each error type targets a distinct failure mode observed in LLM-generated spatial layouts. Crucially, we chain multiple errors together by injecting 2–3 errors per sample to create rejections that appear superficially reasonable but contain compounding issues, closely mimicking how undertrained models fail in practice. The compiler and collision checker automatically verify that rejected samples are indeed invalid, ensuring clean preference labels. Applying Error Chain to our living room scenario yielded 9,000+ high-quality DPO pairs, enabling the model to learn subtle distinctions between valid and invalid spatial reasoning.

  \subsubsection{Three-Stage Training Pipeline}

  We implement a three-stage training pipeline to progressively build model capabilities. For instance, in a living room scenario, \textbf{Pre-training} uses 60,000+ code-only samples to teach \textit{SpatialGrammar} syntax, object vocabularies, and basic spatial patterns through next-token prediction. \textbf{Supervised Fine-Tuning (SFT)} then trains on 2,800+ (instruction, code) pairs so that the model learns to map natural language requests directly to well-structured SpatialGrammar programs. \textbf{Direct Preference Optimization (DPO)} finally refines the model's judgment using 9,000+ preference pairs generated via Error Chain, encouraging it to favor semantically coherent, collision-free layouts over superficially plausible but flawed alternatives; following~\cite{rafailov2023direct}, we apply DPO directly to the SFT checkpoint without a separate reward model. We train a compact 104M-parameter SLM, \textit{SG-Mini}, built upon the minimind-llm architecture, on a mixture of minimind's pre-training, SFT, and DPO datasets and our synthesized \textit{SpatialGrammar} data, with the full pipeline completing in one week on two NVIDIA RTX 4090 GPUs.

  \section{Experiments}
  \label{sec:experiments}

  \begin{figure*}[!t]
      \centering
      \includegraphics[width=0.85\textwidth]{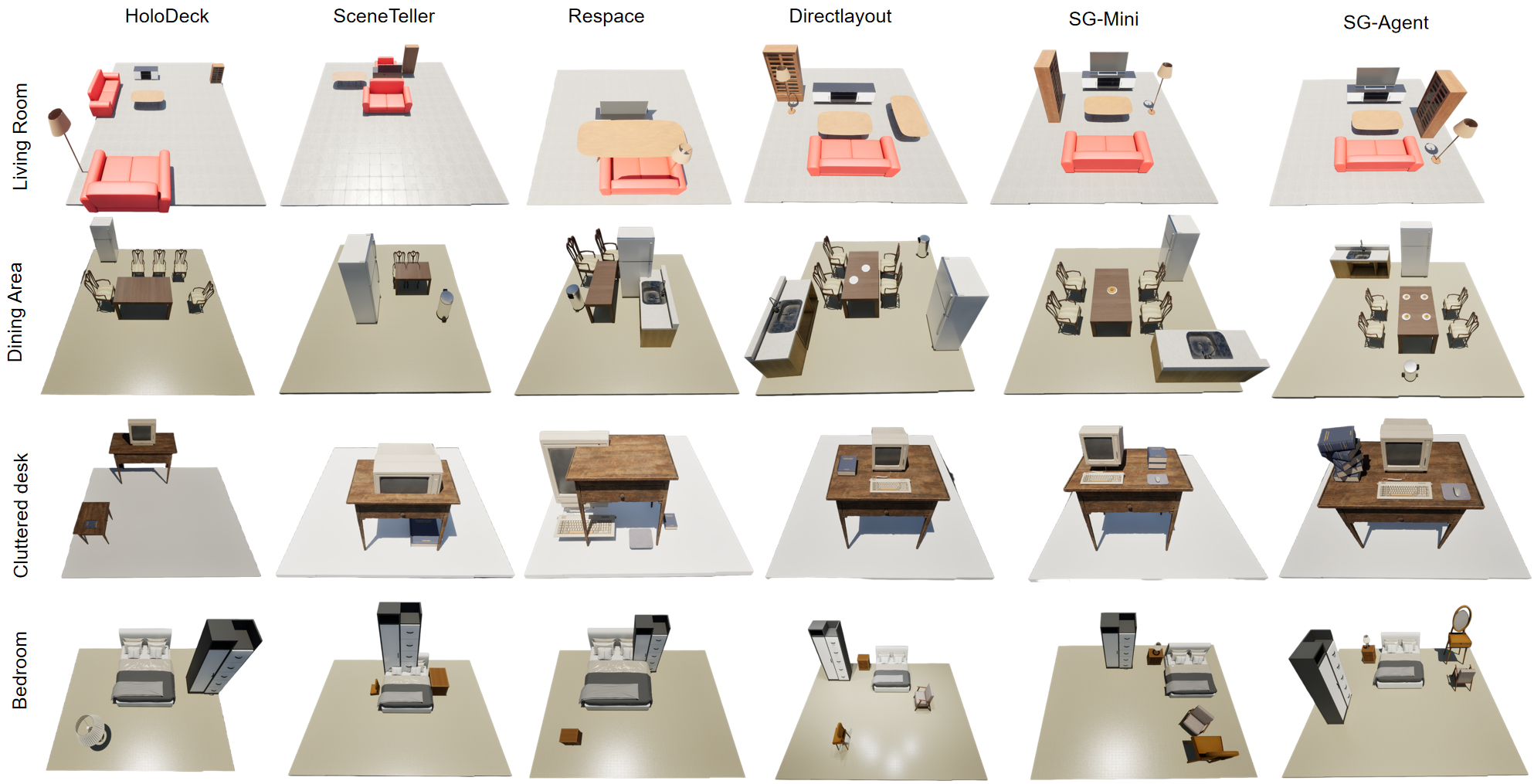}
      \caption{Visual comparison of generation results across different methods in various scenarios. To ensure fairness, we used a unified renderer and consistent assets to render all outputs based on the layout results from these models.}
      \label{fig:results_compare}
      \vspace{-10pt}
  \end{figure*}

  \subsection{Experimental Setup}

  We compare against representative state-of-the-art works covering different technical routes. SG-Mini is our 104M-parameter model trained with the synthetic pipeline from Section~\ref{sec:slm_training}; SG-Agent is the full tool-augmented agent from Section~\ref{sec:agent}, powered by Claude Sonnet 4. We evaluate end-to-end text-to-scene generation across five scenarios of increasing complexity, comprising 159 test scenes (1,198 atomic checks for DRFR): (1)~\textit{single-object placement}, (2)~\textit{multi-object generation}, (3)~\textit{multi-turn editing}, (4)~\textit{hierarchical placement}---objects placed on or inside other objects in parent-local coordinates (e.g., the cluttered desk in Figure~\ref{fig:results_compare}, row 3), and (5)~\textit{architectural generation}. We use five metrics: \textbf{DRFR}~\cite{qin2024infobench}, the fraction of atomic requirements (object presence, spatial relations, constraints) satisfied; \textbf{CR$_{\text{obj}}$ (\%)}, the percentage of objects involved in collisions; \textbf{CLIP Score}~\cite{radford2021learning}, visual--semantic consistency; \textbf{GAS}, Gemini-2.5-Flash aesthetic rating (0--100); and \textbf{HAS}, the mean aesthetic rating from 9 domain professionals on the same 0--100 scale. Our primary claims rest on DRFR and CR$_{\text{obj}}$; GAS and HAS are complementary. All methods receive identical prompts and are evaluated with their complete pipelines (including built-in refinement); SG-Mini operates in single-shot mode. SceneCraft~\cite{hu2024scenecraft} and OptiScene~\cite{yang2025optiscene} are excluded as neither has released reproducible code or weights.

  \subsection{Ablation Studies}

  \textbf{System design ablation} (Table~\ref{tab:ablation_system}). Compared to the same LLM generating raw JSON coordinates without our DSL, introducing SpatialGrammar yields a large improvement in DRFR (0.62$\rightarrow$0.83) and a dramatic reduction in CR$_{\text{obj}}$ (66.7$\rightarrow$13.6). Adding physical and visual feedback yields the best performance (DRFR 0.90, CR$_{\text{obj}}$ 0), indicating that each component contributes meaningfully.

  \begin{table}[h]
  \centering
  \scriptsize
  \setlength{\tabcolsep}{1mm}
  \caption{Ablation on the agentic framework components.}
  \label{tab:ablation_system}
  \begin{tabular}{lcccc}
  \toprule
  \textbf{DSL} & \textbf{Phys. Fbk} & \textbf{Vis. Fbk} & \textbf{DRFR}$\uparrow$ & \textbf{CR$_{\text{obj}}$}$\downarrow$ \\
  \midrule
   & & & 0.62 & 66.7\\
  $\checkmark$ & & & 0.83 & 13.6\\
  $\checkmark$ & $\checkmark$ & & 0.87 & 0 \\
  $\checkmark$ & $\checkmark$ & $\checkmark$ & 0.90 & 0 \\
  \bottomrule
  \end{tabular}
  \vspace{-5pt}
  \end{table}

\begin{figure}[h]
      \centering
      \includegraphics[width=0.9\linewidth]{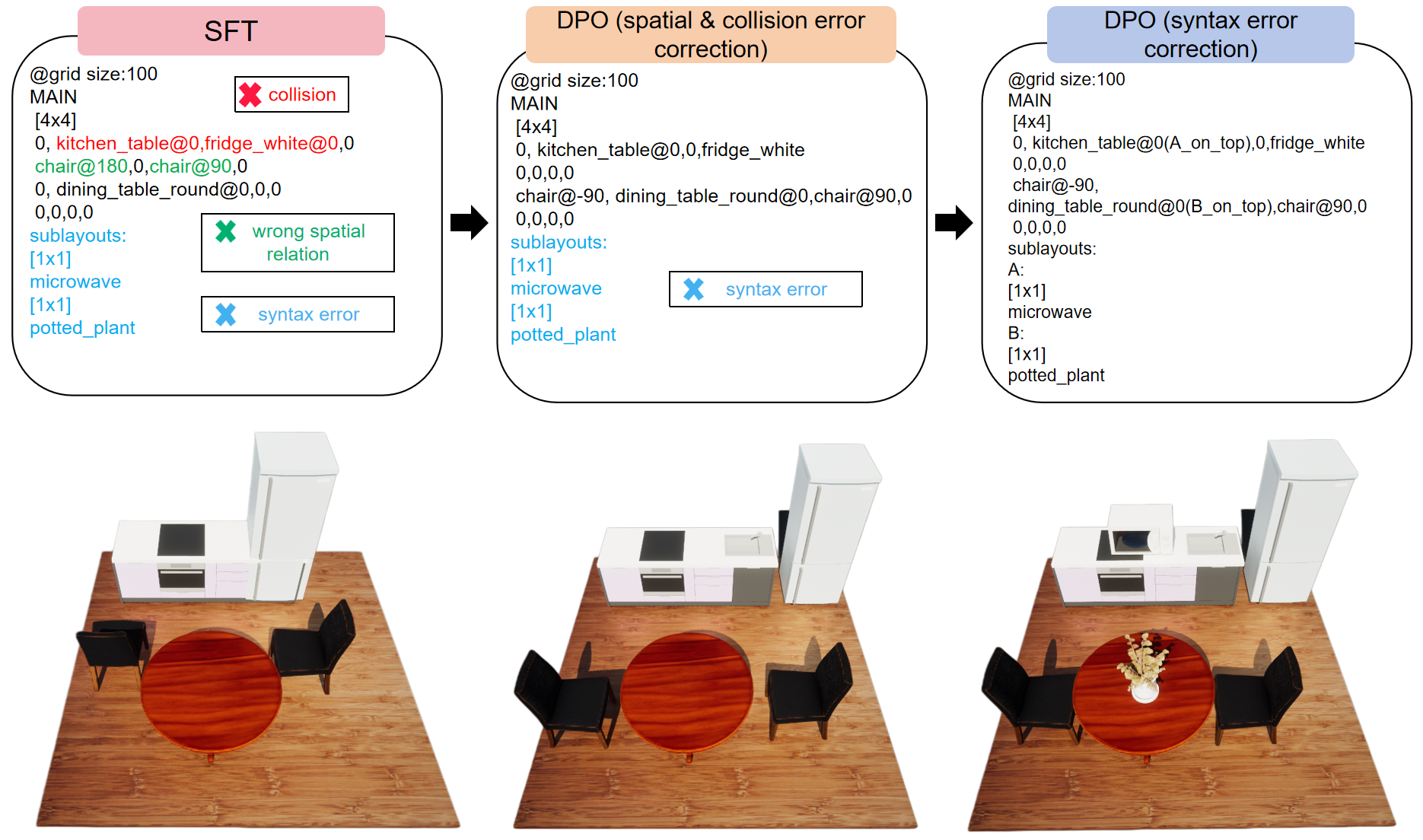}
      \caption{Qualitative example of DPO training effects: injecting different error types progressively corrects collision, spatial, and syntax errors.}
      \label{fig:dpo_effect}
      \vspace{-15pt}
  \end{figure}

  \textbf{Error Chain ablation} (Table~\ref{tab:ablation_dpo}; Figure~\ref{fig:dpo_effect} provides a qualitative example of the progressive correction). We incrementally add each error category to the DPO training set. Collision error injection yields the largest CR$_{\text{obj}}$ improvement (28.4$\rightarrow$16.7). Adding spatial and semantic errors further improves DRFR. Syntax error injection targets code well-formedness and improves DRFR to 0.79 without reducing collision rate, as expected. The full chain achieves the best DRFR, demonstrating that chaining diverse error types produces complementary learning signals.

  \begin{table}[h]
  \centering
  \scriptsize
  \setlength{\tabcolsep}{1.5mm}
  \caption{Ablation on DPO error types. Error types are added incrementally.}
  \label{tab:ablation_dpo}
  \begin{tabular}{lcc}
  \toprule
  \textbf{DPO Error Types} & \textbf{DRFR}$\uparrow$ & \textbf{CR$_{\text{obj}}$}$\downarrow$ \\
  \midrule
  None (SFT only) & 0.55 & 28.4 \\
  + Collision & 0.63 & 16.7 \\
  + Spatial relation & 0.71 & 13.2 \\
  + Semantic & 0.76 & 10.5 \\
  + Syntax (full chain) & \textbf{0.79} & 11.6 \\
  \bottomrule
  \end{tabular}
  \end{table}

  \begin{figure*}[!t]
      \centering
      \includegraphics[width=0.9\textwidth]{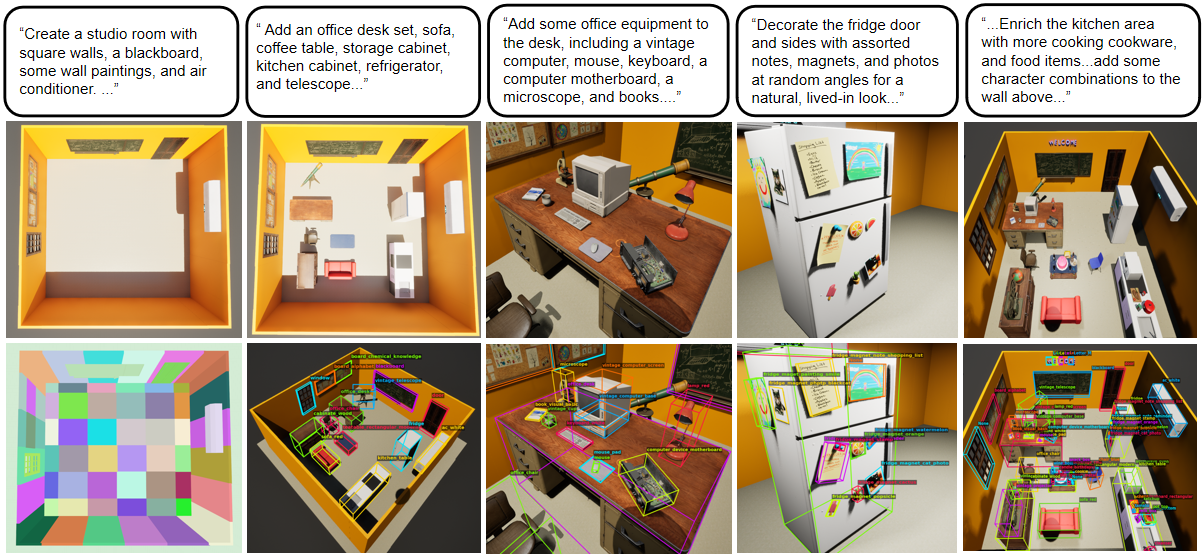}
      \caption{Our Agent system supports contextual memory-based iterative scene refinement, progressively evolving a simple wall structure into a complex 3D box layout and its corresponding indoor scene.}
      \label{fig:iterative_demo}
      \vspace{-10pt}
  \end{figure*}

  \begin{figure}[!t]
      \centering
      \includegraphics[width=0.9\linewidth]{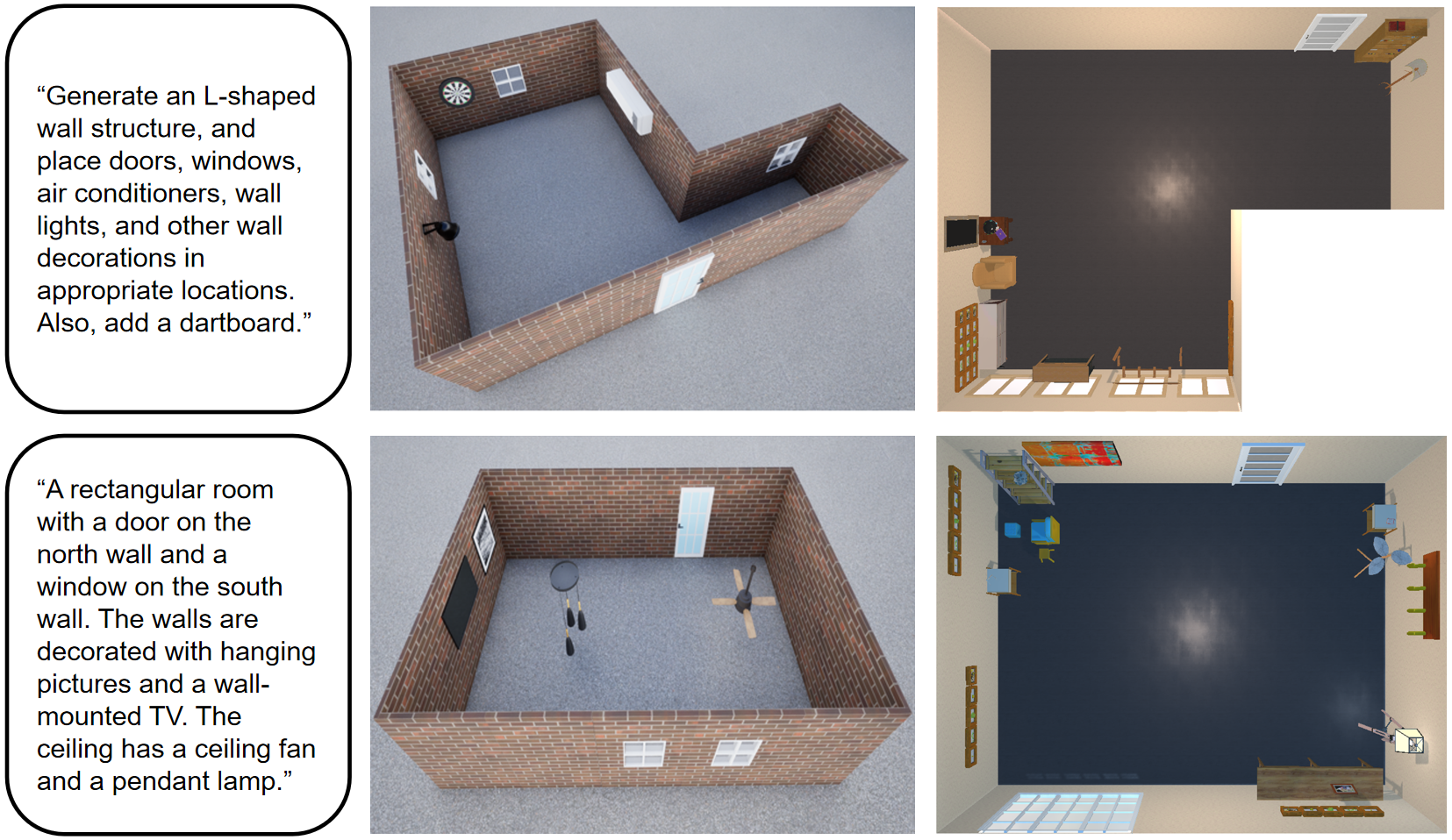}
      \caption{Architectural generation examples showing wall shape control, wall-mounted objects, and ceiling decorations.}
      \label{fig:wall_shape}
      \vspace{-10pt}
  \end{figure}

  \subsection{Quantitative Comparison}
  Table~\ref{tab:main_scenarios} and Figure~\ref{fig:results_compare} report results on three core scenarios, including DirectLayout~\cite{ran2025direct} as a BEV-based baseline. On \textbf{single-object placement}, SG-Mini achieves the highest DRFR (0.99); all three BEV-based methods outperform non-BEV baselines, corroborating that BEV representations benefit spatial reasoning. On \textbf{multi-object generation}, SG-Agent leads (DRFR 0.83); DirectLayout (0.77) surpasses SG-Mini (0.62), reflecting backbone capacity differences rather than a DSL limitation, since SG-Agent on the same DSL achieves the best results. Note that while HoloDeck achieves CR$_{\text{obj}}$=0, its DRFR remains low (0.55), suggesting its constraint solver avoids collisions partly by omitting requested objects. For \textbf{hierarchical placement}, SG-Mini (0.69) outperforms DirectLayout (0.61) across all metrics, and SG-Agent (0.93) leads with zero collisions, demonstrating the advantage of our sub-layout mechanism---absent from DirectLayout's flat BEV. Across all core scenarios, SG-Agent also achieves the highest CLIP scores, indicating strong visual--semantic alignment. GAS and HAS show broadly consistent trends, yet they capture different aspects of generation quality. Human evaluators tend to be positively biased by visually uncrowded and clean layouts (e.g., HoloDeck), even when objects are erroneously omitted. On the other hand, GAS is more anchored to the global text-image semantic alignment rather than pure aesthetic preference, though its evaluation remains noisy due to the limited visual reasoning capacity of VLMs for dense 3D scenes.

  Table~\ref{tab:extended_scenarios} evaluates extended scenarios. On \textbf{multi-turn editing}, SG-Agent achieves DRFR 0.92 with zero collisions, outperforming Respace (0.37). Figure~\ref{fig:iterative_demo} illustrates how the agent progressively builds a complex scene through iterative dialogue. SG-Mini is excluded as this scenario requires agentic capabilities beyond a 104M-parameter model. On \textbf{architectural generation}, both SG-Agent (0.85) and SG-Mini (0.81) outperform HoloDeck (0.40); Figure~\ref{fig:wall_shape} shows examples of non-rectangular wall shapes, wall-mounted objects, and ceiling elements that our DSL can specify but HoloDeck cannot. Overall, SG-Agent delivers the best performance across all scenarios, while SG-Mini remains competitive and excels on hierarchical tasks where our sub-layout design provides a structural advantage.

  \begin{table}[t]
  \centering
  \scriptsize
  \setlength{\tabcolsep}{1.2mm}
  \caption{Results on core generation scenarios. CR$_{\text{obj}}$ is not applicable (---) for single-object placement.}
  \label{tab:main_scenarios}
  \begin{tabular}{llccccc}
  \toprule
  \textbf{Scenario} & \textbf{Method} & \textbf{DRFR}$\uparrow$ & \textbf{CR$_{\text{obj}}$}$\downarrow$ & \textbf{CLIP}$\uparrow$ & \textbf{GAS}$\uparrow$ & \textbf{HAS}$\uparrow$ \\
  \midrule
  \multirow{6}{*}{Single Object} 
  & SceneTeller & 0.62 & --- & 23.98 & 66.25 & 60.22 \\
  & Respace & 0.63 & --- & 24.42 & 39.91 & 33.17 \\
  & HoloDeck & 0.48 & --- & 24.90 & 73.73 & 54.31 \\
  & DirectLayout & 0.92 & --- & 26.23 & 79.04 & 76.41 \\
  & \textbf{SG-Mini} & \textbf{0.99} & --- & 24.46 & 80.71 & 78.56 \\
  & \textbf{SG-Agent} & 0.97 & --- & \textbf{28.42} & \textbf{84.20} & \textbf{79.33} \\
  \midrule
  \multirow{6}{*}{Multi-Object} 
  & SceneTeller & 0.48 & 52.6 & 24.34 & 56.84 & 48.67 \\
  & Respace & 0.42 & 23.7 & 24.81 & 56.71 & 51.22 \\
  & HoloDeck & 0.55 & \textbf{0} & 25.16 & 66.84 & 75.89 \\
  & DirectLayout & 0.77 & 9.93 & 27.23 & 69.91 & 73.5 \\
  & \textbf{SG-Mini} & 0.62 & 11.07 & 27.87 & 67.24 & 60.31 \\
  & \textbf{SG-Agent} & \textbf{0.83} & 3.40 & \textbf{28.20} & \textbf{73.55} & \textbf{77.56} \\
  \midrule
  \multirow{6}{*}{Hierarchical} 
  & HoloDeck & 0.36 & \textbf{0} & 21.69 & 41.76 & 49.33 \\
  & SceneTeller & 0.45 & 31.0 & 20.75 & 31.18 & 24.44 \\
  & Respace & 0.41 & 41.0 & 23.54 & 50.29 & 43.63 \\
  & DirectLayout & 0.61 & 11.65 & 24.95 & 58.23 & 64.71 \\
  & \textbf{SG-Mini} & 0.69 & 3.33 & 26.57 & 68.94 & 65.1 \\
  & \textbf{SG-Agent} & \textbf{0.93} & \textbf{0} & \textbf{31.05} & \textbf{93.24} & \textbf{72.19} \\
  \bottomrule
  \end{tabular}
  \vspace{-5pt}
  \end{table}

  \begin{table}[t]
  \centering
  \scriptsize
  \setlength{\tabcolsep}{1.2mm}
  \caption{Results on extended scenarios: multi-turn editing and architectural generation. CR$_{\text{obj}}$ is not applicable (---) for architectural generation.}
  \label{tab:extended_scenarios}
  \begin{tabular}{llccccc}
  \toprule
  \textbf{Scenario} & \textbf{Method} & \textbf{DRFR}$\uparrow$ & \textbf{CR$_{\text{obj}}$}$\downarrow$ & \textbf{CLIP}$\uparrow$ & \textbf{GAS}$\uparrow$ & \textbf{HAS}$\uparrow$ \\
  \midrule
  \multirow{2}{*}{Multi-turn} 
  & Respace & 0.37 & 50.0 & 26.31 & 47.90 & 42.89 \\
  & \textbf{SG-Agent} & \textbf{0.92} & \textbf{0} & \textbf{30.29} & \textbf{85.36} & \textbf{75.6} \\
  \midrule
  \multirow{3}{*}{Architectural} 
  & HoloDeck & 0.40 & --- & 23.38 & 63.08 & 73.56 \\
  & \textbf{SG-Mini} & 0.81 & --- & 27.12 & 67.53 & 65.78 \\
  & \textbf{SG-Agent} & \textbf{0.85} & --- & \textbf{28.15} & \textbf{80.75} & \textbf{83.12} \\
  \bottomrule
  \end{tabular}
  \vspace{-10pt}
  \end{table}

  \section{Conclusion}

  This paper introduced \textit{SpatialGrammar}, a domain-specific language that represents indoor layouts as BEV grid placements with deterministic compilation to valid 3D geometry. This design encodes physical priors directly into the representation, enabling verifiable constraint checking during generation. Building on this foundation, we developed \textit{SG-Agent}, a closed-loop system that uses compiler feedback to iteratively refine scenes, and \textit{SG-Mini}, a 104M-parameter model trained entirely on compiler-validated synthetic data that achieves competitive performance on core generation scenarios. Code will be released upon acceptance.

\bibliographystyle{ACM-Reference-Format}
\bibliography{main}

\newpage
\appendix
\onecolumn

\begin{figure*}[!t]
    \centering
    \includegraphics[width=0.8\textwidth]{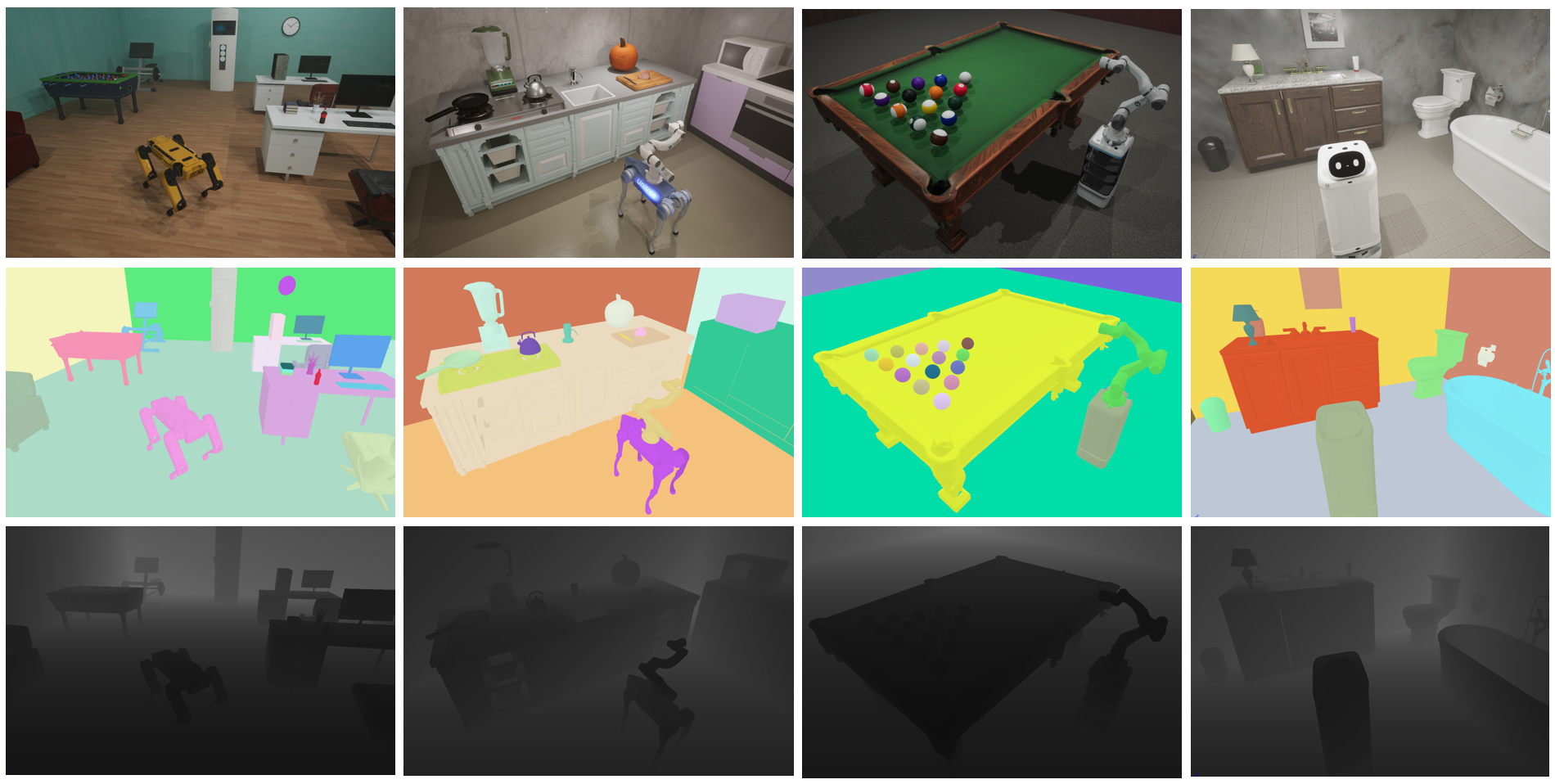}
    \caption{Diverse display of scenes generated by our model and GT of instance segmentation \& depth.}
    \label{fig:visdemoGTs}
    \vspace{-10pt}
\end{figure*}

\section{Evaluation Setup Details}
\label{sec:appendix_benchmark}

This section provides a detailed breakdown of our evaluation setup. We designed five scenarios to systematically evaluate the full range of capabilities needed for 3D scene generation, from basic object placement to complex spatial reasoning. The dataset consists of \textbf{159} distinct test scenes, which are further broken down into \textbf{1,198} specific checkpoints for precise evaluation. Table~\ref{tab:benchmark_statistics} summarizes the statistics and objectives for each task.

\begin{table*}[t]
    \centering
    \small
    \caption{Summary of the Evaluation Scenario Statistics and Objectives.}
    \label{tab:benchmark_statistics}
    \begin{tabular}{p{0.25\textwidth} c c p{0.45\textwidth}}
    \toprule
    \textbf{Task Category} & \textbf{Scenes} & \textbf{Checks} & \textbf{Core Objective} \\
    \midrule
    Task 1: Single Object & 56 & 213 & Basic placement \& orientation correctness in global coordinates. \\
    Task 2: Multiple Objects & 38 & 656 & 3D spatial planning, relative positioning, and collision avoidance. \\
    Task 3: Conversational & 28 & 74 & Context maintenance and incremental updates across dialogue turns. \\
    Task 4: Hierarchical Placement & 17 & 193 & Hierarchical reasoning using local coordinate systems (e.g., on/in). \\
    Task 5: Architectural & 20 & 62 & Structural validity (closed loops) and metric scale of room elements. \\
    \midrule
    \textbf{Total} & \textbf{159} & \textbf{1,198} & \\
    \bottomrule
    \end{tabular}
\end{table*}

\subsection{Test Objective}

Our evaluation setup is structured to progressively evaluate the model's spatial intelligence, starting from fundamental coordinate understanding and advancing to complex structural and interactive reasoning.

\textbf{Task 1 (Single Object Placement)} serves as the foundational test for any 3D generative model. It rigorously verifies whether the model has mastered the global coordinate system and absolute cardinal directions, which are prerequisites for more complex layouts. For instance, a prompt like \textit{``Place an armchair in the center of the room, facing west''} directly tests the model's ability to map semantic location and orientation terms to precise geometric transformations.

Building on this foundation, \textbf{Task 2 (Multiple Object Combination)} introduces the challenge of 3D spatial planning and collision avoidance. It requires the model to understand relative spatial descriptions (e.g., ``left of'', ``opposite'') and manage the distribution of multiple objects within a shared space. A typical prompt, such as \textit{``Place a TV stand... with a TV on top... Place a sofa... facing the TV...''}, demands that the model not only places individual items correctly but also maintains logical functional relationships between them (e.g., sofa facing the TV, coffee table in front of the sofa) without physical overlap.

\textbf{Task 3 (Multi-turn Conversational Editing)} shifts the focus to temporal consistency and context maintenance, which are critical for interactive design agents. It evaluates the model's ability to perform incremental updates—such as adding a wardrobe or moving a chair—based on a sequence of user instructions (e.g., \textit{User: ``Add a wardrobe'' $\rightarrow$ User: ``Add a sofa''}), ensuring that new changes do not disrupt the previously established scene state.

Real-world scenes are characterized by rich hierarchical structures, which we evaluate in \textbf{Task 4 (Hierarchical Placement)}. This task assesses the model's capability to handle parent-child dependencies using local coordinate systems, rather than placing everything in global coordinates. Complex instructions like \textit{``Generate a dining table with a stack of books... arranged in a spiral pattern''} test the model's fine-grained control over sub-layouts and object attachments.

Finally, \textbf{Task 5 (Architectural Generation)} examines the model's understanding of the structural container itself. Unlike object placement, generating architecture requires topological reasoning (e.g., ensuring walls form closed loops) and metric scaling. Prompts such as \textit{``Generate a T-shaped wall structure with a window on each... segment''} verify that the model can construct valid, physically plausible room environments that can house the generated furniture.

\section{Evaluation Process}
\label{sec:appendix_metrics}

\subsection{Decomposed Requirements Following Ratio}

The \textbf{Decomposed Requirements Following Ratio (DRFR)} is a fine-grained metric designed to rigorously evaluate the instruction-following capabilities of generative models. Unlike holistic scoring (which is subjective) or CLIP scores (which capture global semantics but miss fine-grained spatial details), DRFR decomposes a complex natural language instruction into a set of \textbf{atomic}, \textbf{verifiable}, and \textbf{binary (Yes/No)} constraints.

\subsubsection{Inspection Details}
In the context of 3D scene generation, we adapted the DRFR metric to specifically target spatial and structural correctness. Our decomposition taxonomy includes four primary categories:
\begin{itemize}
    \item \textbf{Existence Check ($\mathcal{C}_{exist}$):} Verifies whether the objects mentioned in the prompt are present in the scene (e.g., \textit{``Is there a sofa?''}).
    \item \textbf{Attribute Check ($\mathcal{C}_{attr}$):} Checks for correct visual attributes such as color, texture, or shape (e.g., \textit{``Is the table round?''}).
    \item \textbf{Spatial Relation Check ($\mathcal{C}_{spatial}$):} Evaluates the relative positioning and global layout correctness (e.g., \textit{``Is the lamp to the left of the bed?''}).
    \item \textbf{Hierarchical/Support Check ($\mathcal{C}_{hier}$):} Verifies physical dependencies and attachment relations, crucial for our sub-layout system (e.g., \textit{``Is the vase physically supported by the table?''}).
\end{itemize}

The final score is calculated as the ratio of satisfied atomic requirements to the total number of requirements:
\begin{equation}
    \text{DRFR} = \frac{\sum \mathbb{I}(\text{requirement satisfied})}{\text{Total Requirements}}
\end{equation}

\subsubsection{Cumulative Evaluation for Conversational Editing (Task 3)}
\textbf{Task 3 (Multi-turn Conversational Editing)} presents a unique challenge: evaluating the consistency of a scene that evolves over time. A single-step evaluation is insufficient because the model must not only execute the current instruction (e.g., \textit{``Add a chair''}) but also maintain the state established by all previous turns (e.g., keeping the previously generated \textit{TV stand} and \textit{sofa} intact).

Therefore, we adopt a \textbf{Cumulative State Decomposition} strategy for Task 3. For each turn $t$ in the dialogue, the set of atomic requirements $R_t$ includes:
\begin{itemize}
    \item Constraints from the current instruction $I_t$.
    \item All persistent constraints from history $H_{t-1} = \{I_1, ..., I_{t-1}\}$ that have not been explicitly modified or removed.
\end{itemize}
For example, if Turn 1 requests a ``TV stand'' and Turn 2 requests an ``armchair'', the evaluation at Turn 2 checks for the existence and correct placement of \textit{both} the TV stand and the armchair. This ensures that the metric captures phenomena like catastrophic forgetting or accidental deletion of objects. Figure~\ref{fig:drfr_task3} visually illustrates this cumulative check process, where the checklist grows as the conversation progresses.

\subsubsection{Visual Demonstrations}
\begin{figure}[h]
    \centering
    \includegraphics[width=0.8\linewidth]{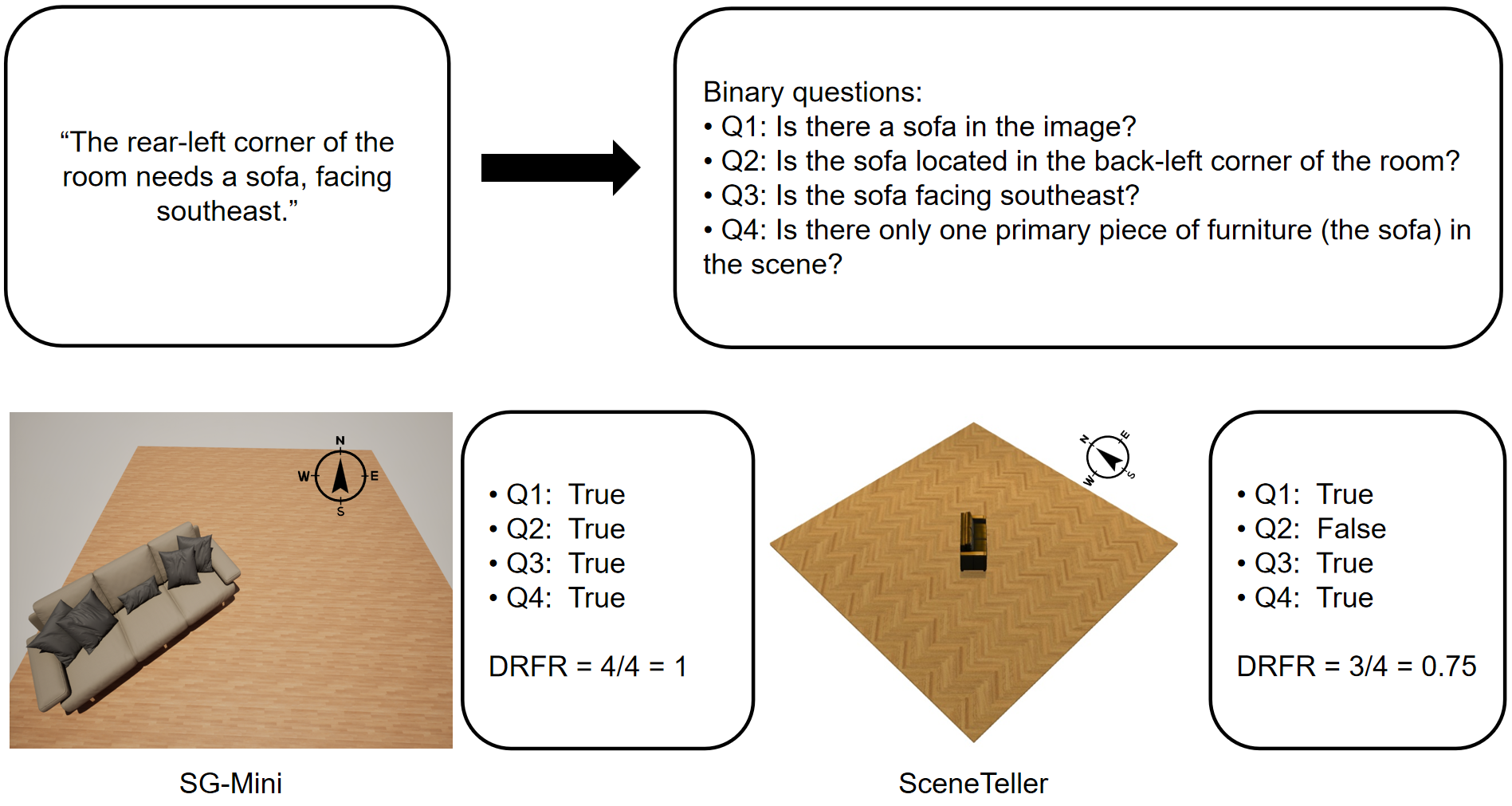}
    \caption{DRFR evaluation for \textbf{Task 1 (Single Object Placement)}. Visual comparison between SG-Mini and SceneTeller, showing the decomposition of atomic requirements and their verification.}
    \label{fig:drfr_task1}
\end{figure}

\begin{figure}[h]
    \centering
    \includegraphics[width=0.8\linewidth]{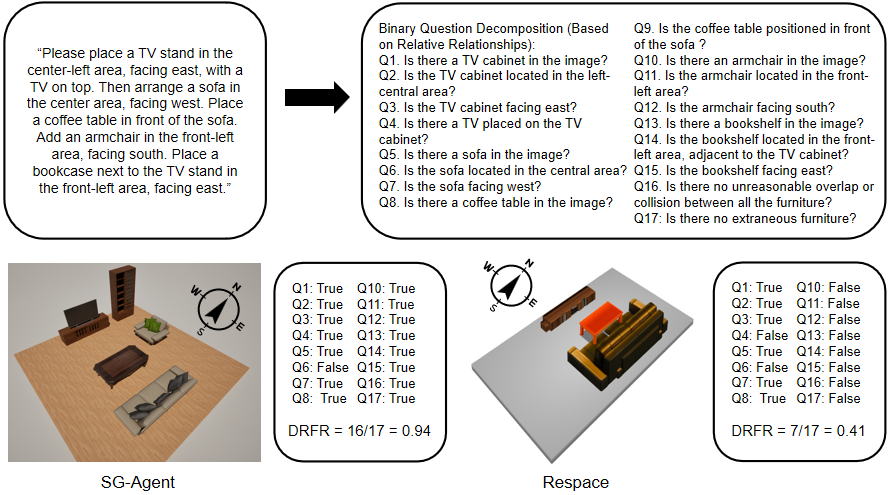}
    \caption{DRFR evaluation for \textbf{Task 2 (Multiple Object Combination)}. Visual comparison between SG-Agent and Respace on a complex layout task, highlighting the detailed spatial checks.}
    \label{fig:drfr_task2}
\end{figure}

\begin{figure}[h]
    \centering
    \includegraphics[width=0.8\linewidth]{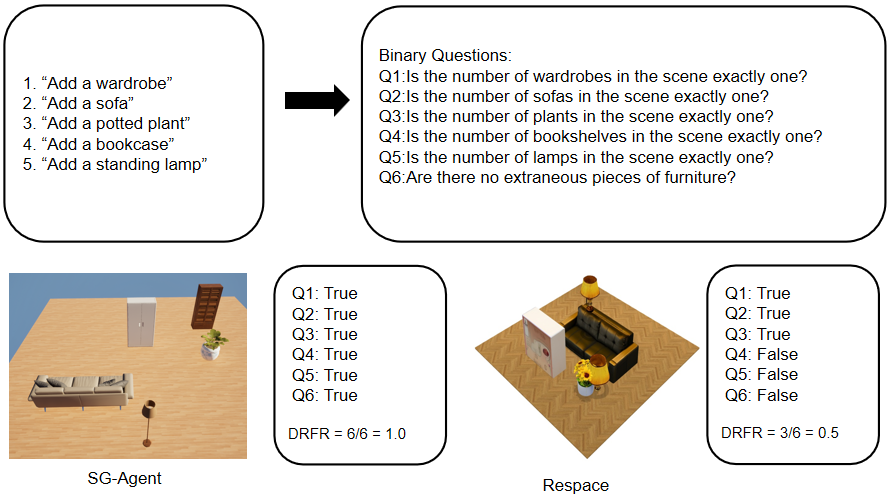}
    \caption{DRFR evaluation for \textbf{Task 3 (Multi-turn Conversational Editing)}. This example validates whether incremental updates (e.g., adding objects) are correctly executed without disrupting the existing scene state. The checklist reflects the cumulative state of the scene.}
    \label{fig:drfr_task3}
\end{figure}

\begin{figure}[h]
    \centering
    \includegraphics[width=0.8\linewidth]{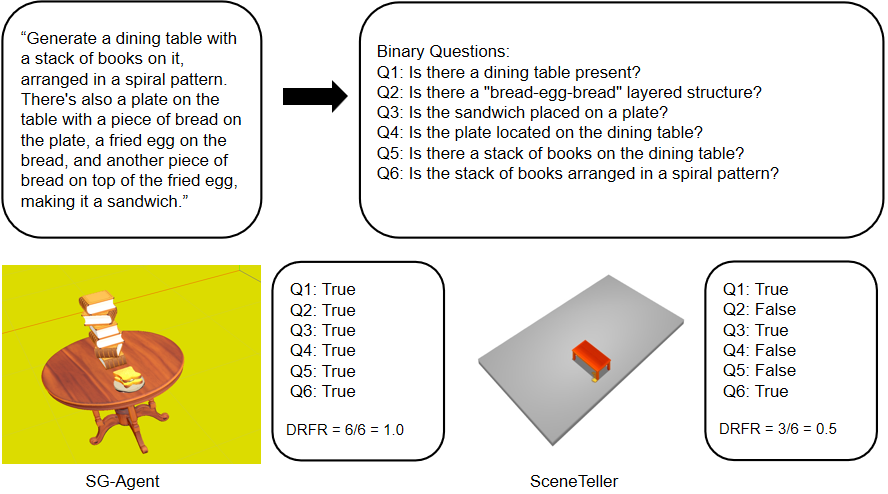}
    \caption{DRFR evaluation for \textbf{Task 4 (Hierarchical Placement)}. The evaluation focuses on verifying hierarchical constraints (e.g., ``on top of'') and the precision of local coordinate systems.}
    \label{fig:drfr_task4}
\end{figure}

\begin{figure}[h]
    \centering
    \includegraphics[width=0.8\linewidth]{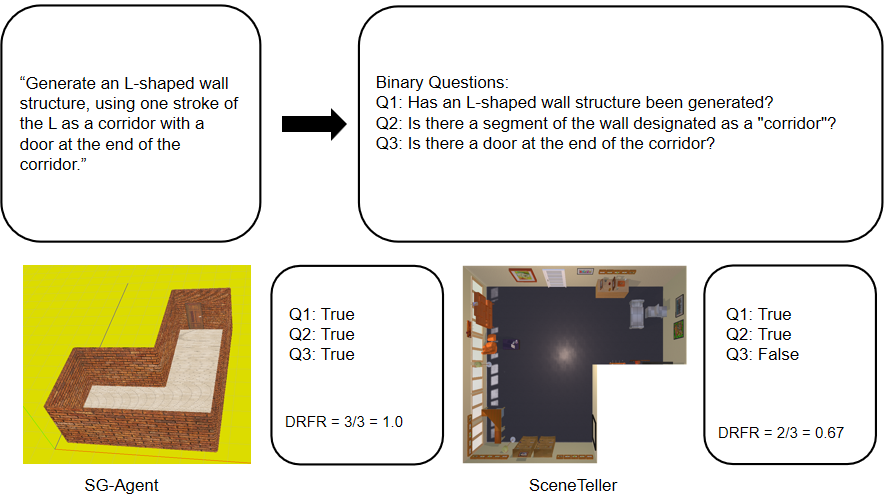}
    \caption{DRFR evaluation for \textbf{Task 5 (Architectural Generation)}. The checkpoints verify structural integrity (e.g., closed loops) and the presence/placement of architectural features like doors and windows.}
    \label{fig:drfr_task5}
\end{figure}

Figures~\ref{fig:drfr_task1} to \ref{fig:drfr_task5} provide detailed visual breakdowns of how instructions are decomposed and evaluated across all five tasks.

\subsection{Gemini Aesthetic Score (GAS)}

While DRFR focuses on objective instruction following, we employ the \textbf{Gemini Aesthetic Score (GAS)} to assess the visual quality, physical plausibility, and overall rationality of the generated scenes.

To ensure full reproducibility, we provide the complete system prompts used for each task. These prompts were fed into the Gemini model to generate the scores.

\subsubsection{Task 1: Single Object Placement}
\begin{quote}
\scriptsize
\textbf{System Prompt:} You are a professional 3D scene evaluation expert. Please evaluate this rendered image according to the following standards:

\textbf{Important Note}: Since the rendering viewpoint is not standardized, focus on \textbf{relative positioning and orientation logic} rather than absolute accuracy. Evaluate whether the furniture's position and orientation are \textbf{internally consistent} with the described location.

\textbf{Evaluation Criteria}:
\begin{enumerate}
    \item \textbf{Furniture Existence \& Type Accuracy (35 points)}: Exactly the right furniture type as requested (35 pts); Very similar furniture type (25-30 pts); ... Penalty for Extra Furniture (-20 pts).
    \item \textbf{Relative Position Logic (30 points)}: Position perfectly logical relative to room layout (30 pts); Position mostly logical (20-25 pts); ...
    \item \textbf{Relative Orientation Logic (25 points)}: Orientation perfectly matches description (25 pts); Orientation mostly correct (18-22 pts); ...
    \item \textbf{Scene Rationality (10 points)}: Overall scene composition is natural and realistic.
\end{enumerate}
\end{quote}

\subsubsection{Task 2: Multiple Object Combination}
\begin{quote}
\scriptsize
\textbf{System Prompt:} You are a professional 3D scene evaluation expert. Please evaluate this rendered image according to the following standards:

\textbf{Important Note}: Task2 focuses on multi-furniture placement. Prioritize instruction following accuracy and collision detection over perfect spatial relationships.

\textbf{Evaluation Criteria}:
\begin{enumerate}
    \item \textbf{Instruction Following Accuracy (40 points)}: All instructions perfectly followed (40 pts); Most instructions followed (30-35 pts); ... Penalty for Extra Furniture (-15 pts).
    \item \textbf{Collision \& Overlap Detection (30 points)}: No furniture collisions (30 pts); Minor overlaps (22-25 pts); Significant overlaps (8-12 pts); Severe collisions (0-5 pts).
    \item \textbf{Overall Layout Rationality (20 points)}: Layout makes practical sense for real-world use.
    \item \textbf{Visual Harmony \& Aesthetics (10 points)}: Scene is visually pleasing and well-balanced.
\end{enumerate}
\end{quote}

\subsubsection{Task 3: Multi-turn Conversational Editing}
\begin{quote}
\scriptsize
\textbf{System Prompt:} You are a professional 3D scene evaluation expert. Please evaluate this rendered image according to the following standards. This image represents the result after executing a sequence of editing instructions. You need to evaluate whether the current scene correctly reflects ALL cumulative edits.

\textbf{Evaluation Criteria}:
\begin{enumerate}
    \item \textbf{Cumulative Instruction Execution (45 points)}: Does the current scene correctly reflect ALL editing operations from step 1 to current step? All executed (45 pts); Most executed (35-40 pts); ...
    \item \textbf{Current Step Accuracy (25 points)}: Is the current step instruction (ADD/REMOVE/REPLACE) correctly executed? Perfectly executed (25 pts); Mostly executed (18-22 pts); ...
    \item \textbf{Scene Consistency \& Logic (20 points)}: Scene is internally consistent and logical.
    \item \textbf{Edit Feasibility \& Realism (10 points)}: All edits result in realistic, feasible arrangements.
\end{enumerate}
\end{quote}

\subsubsection{Task 4: Hierarchical Placement}
\begin{quote}
\scriptsize
\textbf{System Prompt:} You are a professional 3D scene evaluation expert. Please evaluate this rendered image according to the following standards:

\textbf{Important Note}: The key challenge is correctly handling \textbf{relative spatial relationships} between objects, especially "on top of", "inside", "beside".

\textbf{Evaluation Criteria}:
\begin{enumerate}
    \item \textbf{Spatial Relationship Accuracy (40 points)}: All spatial relationships perfectly executed (40 pts); Most correct (30-35 pts); ...
    \item \textbf{Surface Contact \& Precision (30 points)}: Perfect surface contact with no gaps/overlaps (30 pts); Good contact (22-25 pts); Objects floating or sinking (0-5 pts).
    \item \textbf{Nesting Logic Correctness (20 points)}: Do nested arrangements make physical and logical sense?
    \item \textbf{Overall Layout Rationality (10 points)}: Layout practical and realistic for real-world use.
\end{enumerate}
\end{quote}

\subsubsection{Task 5: Architectural Generation}
\begin{quote}
\scriptsize
\textbf{System Prompt:} You are a professional architecture and structural engineering evaluation expert. Please evaluate this rendered image according to the following standards:

\textbf{Important Note}: Task5 focuses on \textbf{wall structure generation}. Evaluate whether the generated structure accurately follows the architectural specifications.

\textbf{Evaluation Criteria}:
\begin{enumerate}
    \item \textbf{Structural Element Accuracy (45 points)}: Are the specified structural elements (walls, doors, windows) correctly generated? All match (45 pts); Most correct (35-40 pts); ...
    \item \textbf{Geometric Precision (25 points)}: Are the geometric specifications (dimensions, shapes) accurate? Perfect accuracy (25 pts); Good accuracy (18-22 pts); ...
    \item \textbf{Architectural Feasibility (20 points)}: Is the structure buildable and follows basic architectural principles?
    \item \textbf{Spatial Completeness (10 points)}: Does the structure create a complete, enclosed space as intended?
\end{enumerate}
\end{quote}

\subsection{CLIP Score Protocol}

The \textbf{CLIP Score} measures the semantic consistency between the generated image and the text description. We utilize the OpenAI \textbf{CLIP ViT-B/16} model for this evaluation.

\subsubsection{Cumulative State for Conversational Tasks}
Similar to the DRFR evaluation, computing CLIP scores for \textbf{Task 3 (Multi-turn Conversational Editing)} requires special handling. Since CLIP evaluates the alignment between a static image and a text description, using incremental instructions (e.g., \textit{``Add a sofa''}) is inappropriate for assessing the final rendered scene.

Therefore, we construct Cumulative State Descriptions for each turn in Task 3. For instance, if the history involves adding a TV stand followed by a sofa, the prompt used for CLIP calculation at the second turn would be \textit{``A living room with a TV stand and a sofa''}, rather than just the incremental instruction. This ensures that the CLIP score measures the global semantic correctness of the evolving scene.

\subsubsection{Fair Evaluation Strategy}
To ensure a fair comparison across models with varying success rates, we adopt a strict penalty mechanism. If a model fails to generate a scene for a given prompt (resulting in a missing image), the CLIP score for that instance is recorded as \textbf{0}. This prevents models from achieving artificially high average scores by only generating simple scenes and failing on complex ones.

\section{Training Details}
\label{sec:appendix_training}

This section provides further details on how we trained our SG-Mini model and verifies the impact of our training strategy.

\subsection{Hyperparameters}

We trained our SG-Mini model, which is based on the MiniMind architecture (Small, 104M parameters), using the AdamW optimizer on 2x NVIDIA RTX 4090 GPUs throughout all stages.

During the \textbf{Pre-training stage}, we follow a three-phase curriculum (detailed in \S\ref{sec:appendix_training_dynamics}): first on general corpus only, then progressively adding LLMSLI and LLMSLB synthetic data. The model was trained for 10 epochs in total, with a batch size of 32 and a learning rate of $5 \times 10^{-4}$. The maximum sequence length was set to 1539 to accommodate long dependency chains in our SpatialGrammar DSL.

For the \textbf{Supervised Fine-Tuning (SFT) stage}, we adopted a two-phase training strategy to balance general linguistic capabilities with specific spatial instruction following. In the first phase (200 epochs), we employed a \textbf{mixed data strategy}, blending our synthetically generated SpatialGrammar instructions with the general-purpose SFT dataset from MiniMind. In the second phase (100 epochs), we fine-tuned the model exclusively on the high-quality LLMSLI synthetic data to further sharpen its spatial reasoning skills. The batch size was maintained at 32, with a reduced learning rate of $5 \times 10^{-5}$.

\subsection{Training Dynamics}
\label{sec:appendix_training_dynamics}

\begin{figure}[h]
    \centering
    \includegraphics[width=0.8\linewidth]{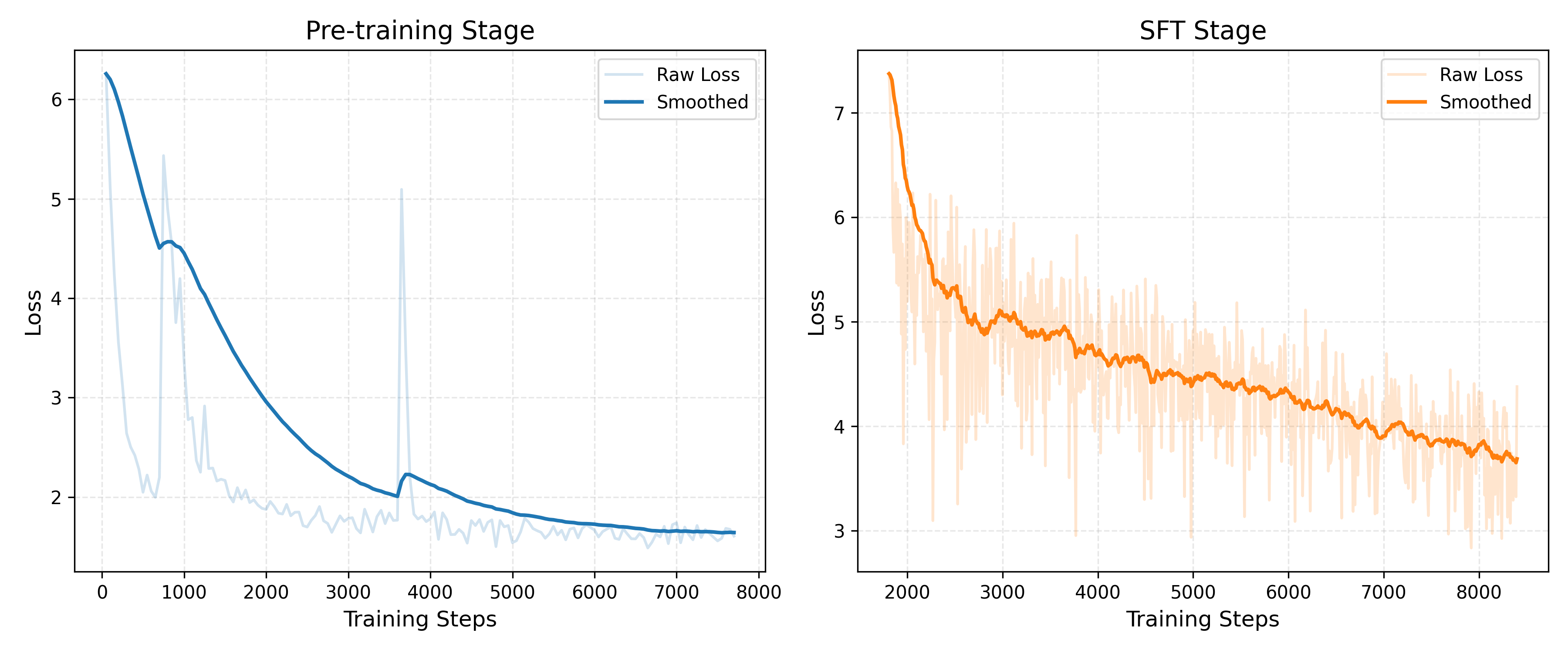}
    \caption{Training loss curves for the Pre-training (left) and SFT (right) stages.}
    \label{fig:loss_curves}
\end{figure}

Figure~\ref{fig:loss_curves} shows the training loss curves during the Pre-training and SFT stages. \textbf{Pre-training} follows a three-phase curriculum strategy: (1) general corpus only, (2) adding LLMSLI synthetic data, and (3) adding LLMSLB synthetic data. The two transient peaks visible in the raw pre-training curve correspond to the introduction of each new domain dataset, where the model temporarily increases loss as it adapts to the new distribution before re-converging. This is expected behavior under curriculum learning and indicates that the model successfully integrates the new data without catastrophic forgetting of previously learned patterns. The \textbf{SFT} loss shows stable convergence throughout the two-phase training (mixed-data phase followed by domain-specific phase), demonstrating successful specialization on SpatialGrammar instruction following.

\section{Grid Size Ablation Study}
\label{sec:appendix_grid_ablation}

The grid cell size is a fundamental hyperparameter in SpatialGrammar that directly trades off spatial precision against token efficiency. We conduct an ablation study across five grid sizes (50cm, 75cm, 100cm, 150cm, 200cm) on a fixed 6m$\times$6m floor area with 9 scenes of varying complexity.

Figure~\ref{fig:grid_ablation_appendix} and Table~\ref{tab:grid_ablation_appendix} summarize the results. DRFR exhibits an inverted-U relationship with grid size, peaking at 100cm: finer grids (50cm) suffer from increased collisions due to the difficulty of placing objects with non-trivial footprints in adjacent cells, while coarser grids (200cm) lack sufficient resolution for complex layouts. The collision rate (CR$_{\text{obj}}$) decreases monotonically with grid size, from 54.9\% at 50cm to 4.9\% at 150cm and beyond. Token count decreases substantially from 152 (50cm) to 18 (200cm), demonstrating the representation's token efficiency. The 100cm default provides a good balance between DRFR performance and collision avoidance, and the configurable grid size allows the model to adapt resolution to scene requirements when needed.

\vspace{0.8em}
\noindent
\begin{minipage}[c]{0.48\textwidth}
    \centering
    \includegraphics[width=\textwidth]{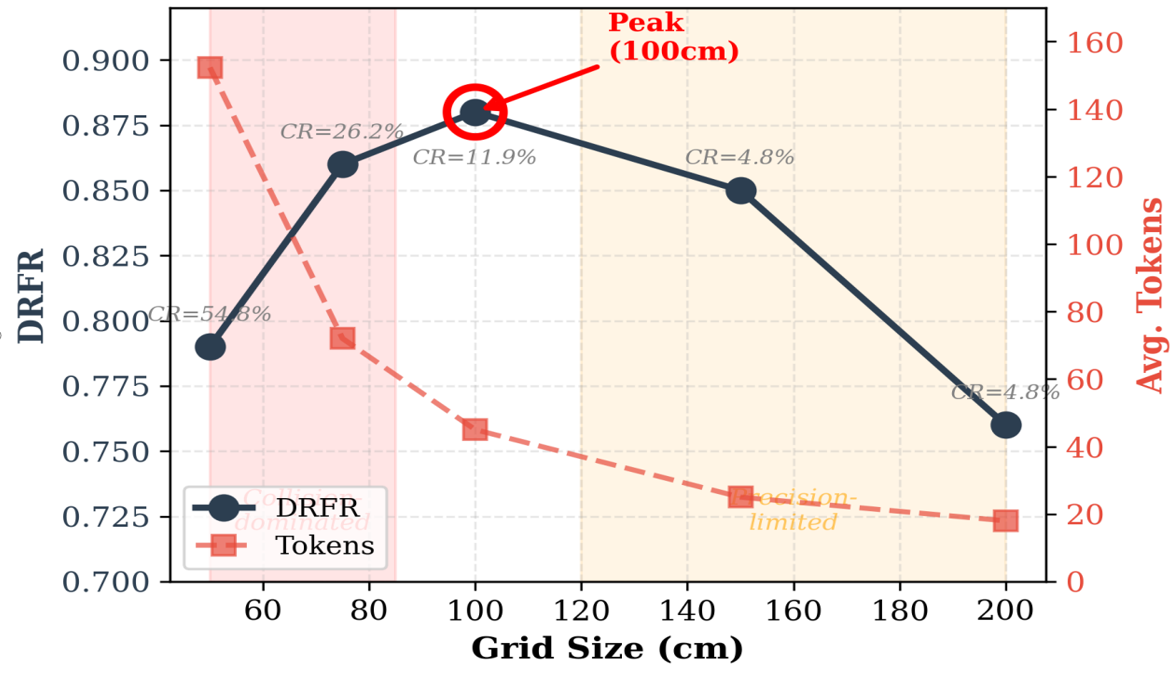}
    \captionof{figure}{Grid size ablation results. DRFR (solid line, left axis) exhibits an inverted-U relationship with grid size, peaking at 100cm. Token count (dashed line, right axis) decreases as grid size increases. Point labels show collision rates (CR$_{\text{obj}}$).}
    \label{fig:grid_ablation_appendix}
\end{minipage}
\hfill
\begin{minipage}[c]{0.48\textwidth}
    \centering
    \small
    \setlength{\tabcolsep}{1.5mm}
    \captionof{table}{Effect of grid cell size on generation quality and token efficiency. All experiments use a fixed 6m$\times$6m floor area across 9 scenes.}
    \label{tab:grid_ablation_appendix}
    \begin{tabular}{ccccc}
    \toprule
    \textbf{Grid Size (cm)} & \textbf{Matrix} & \textbf{DRFR}$\uparrow$ & \textbf{CR$_{\text{obj}}$}$\downarrow$ & \textbf{Avg. Tokens}$\downarrow$ \\
    \midrule
    50  & 12$\times$12 & 0.79 & 54.9\% & 152.3 \\
    75  & 8$\times$8   & 0.86 & 26.2\% & 72.0 \\
    \textbf{100} & \textbf{6$\times$6} & \textbf{0.88} & 11.9\% & 45.0 \\
    150 & 4$\times$4   & 0.85 & \textbf{4.9\%} & 25.0 \\
    200 & 3$\times$3   & 0.76 & 4.9\% & \textbf{18.0} \\
    \bottomrule
    \end{tabular}
\end{minipage}
\vspace{0.8em}

\section{Qualitative Demonstrations}
\label{sec:appendix_qualitative}

We provide additional visual demonstrations of our model's capabilities across two representative scenarios: architectural generation and multi-turn conversational editing.

\subsection{Architectural Generation}

\begin{figure*}[t]
    \centering
    \includegraphics[width=0.8\textwidth]{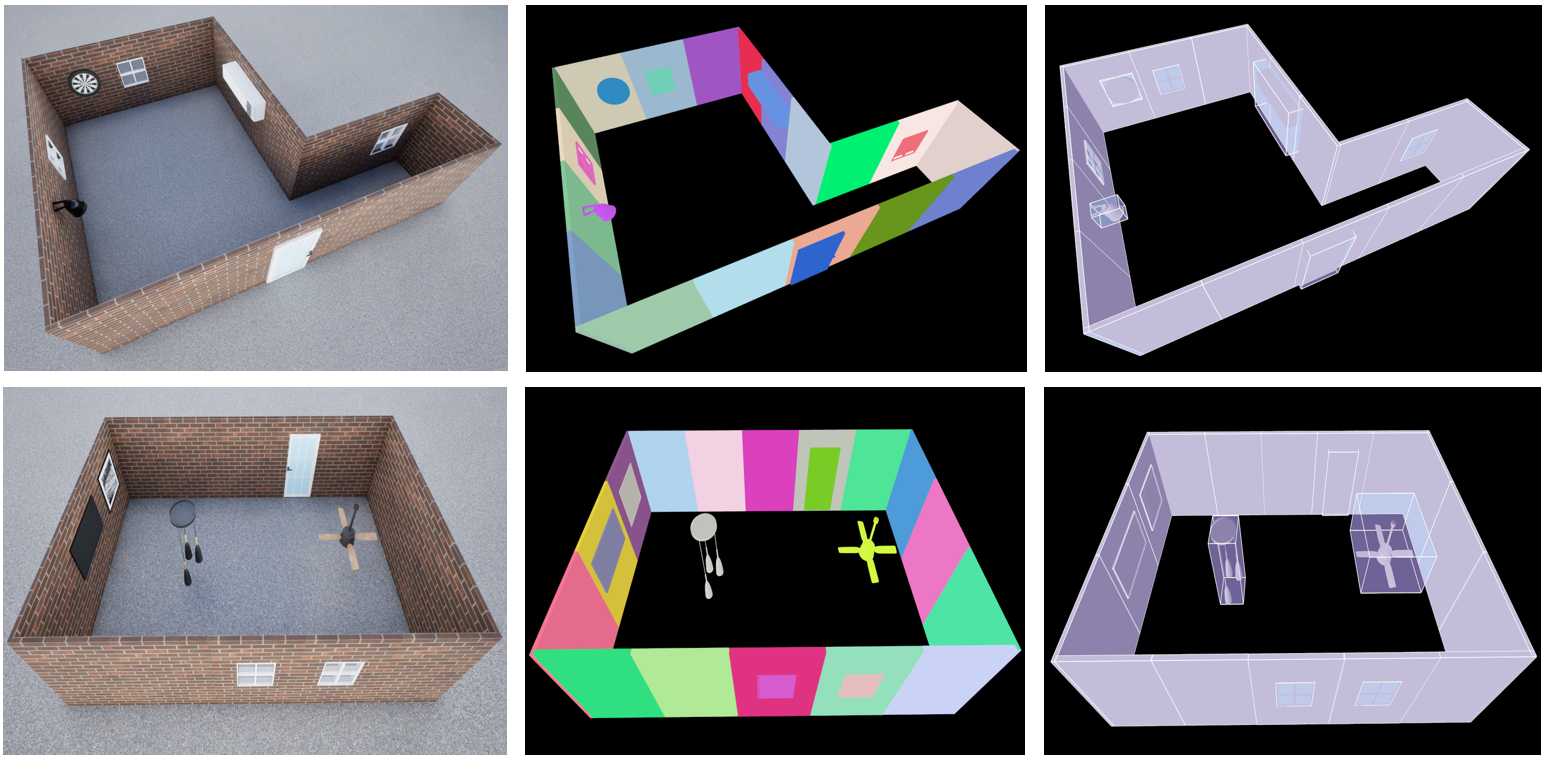}
    \caption{Architectural generation results produced by SG-Agent using the LLMSLB DSL.}
    \label{fig:building_generation}
    \vspace{-15pt}
\end{figure*}

Figure~\ref{fig:building_generation} illustrates the architectural generation capability of our system using the LLMSLB DSL. Each row shows a complete generation pipeline for one building scenario. The left column shows the final high-fidelity render in Unreal Engine 5; the middle column shows the semantic intermediate representation produced by the compiler, where each wall segment and opening is color-coded for analysis; the right column shows the Draft Engine wireframe view used for collision checking and scene validation before asset retrieval. The top row demonstrates generation of a non-rectangular (L-shaped) room structure with correctly placed doors and windows, while the bottom row shows a rectangular room with wall-mounted and ceiling-attached objects (picture frames, mirror, pendant light, ceiling fan), showcasing the face-anchored sub-layout mechanism of LLMSLB.

\subsection{Multi-turn Conversational Editing}

\begin{figure*}[t]
    \centering
    \includegraphics[width=0.8\textwidth]{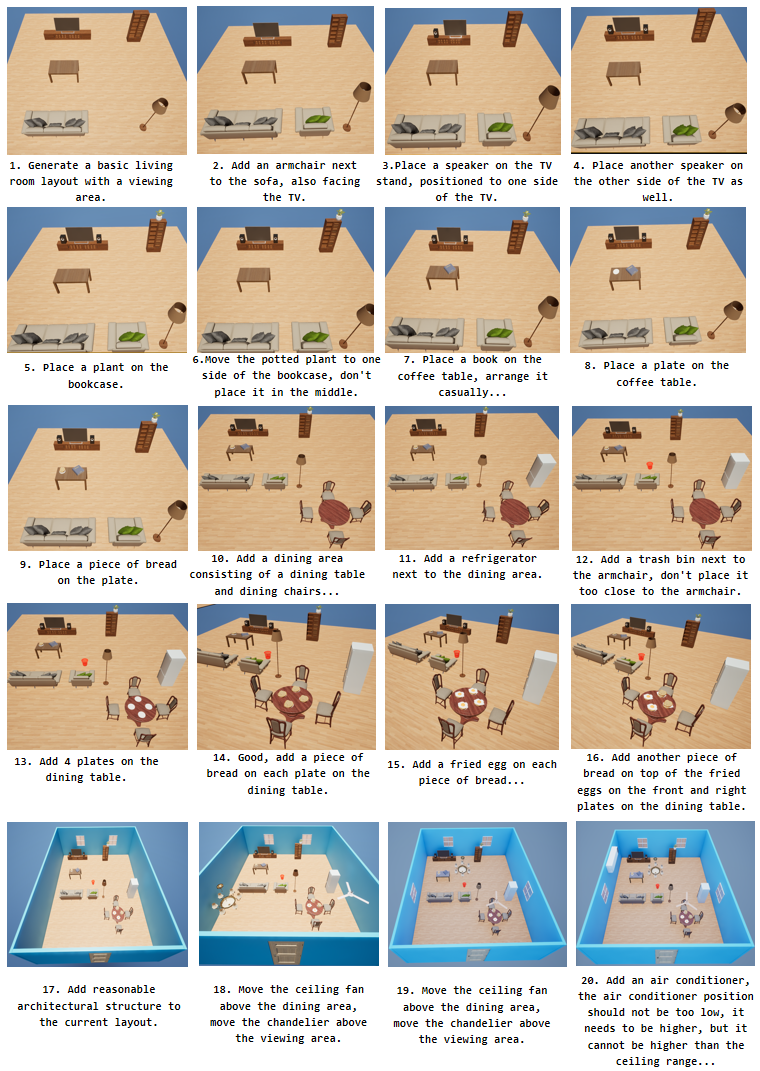}
    \caption{Demonstration of \textbf{Multi-turn Conversational Editing (Task 3)}. The figure presents a sequential visualization of a scene being iteratively refined through natural language dialogue. Each frame corresponds to a specific turn in the conversation, illustrating the model's ability to understand context, perform incremental updates, and maintain global consistency throughout the editing session.}
    \label{fig:multi_turn_example}
\end{figure*}

Figure~\ref{fig:multi_turn_example} showcases the continuous conversational editing capability of our system. It displays a sequence of scene states generated through a multi-turn dialogue, demonstrating how the agent incrementally modifies the scene (e.g., adding objects, changing layouts) based on user instructions while maintaining the context and stability of existing elements.

\end{document}